\documentclass{article}

\usepackage{arxiv}

\usepackage[utf8]{inputenc} 
\usepackage[T1]{fontenc}    
\usepackage{hyperref}       
\usepackage{url}            
\usepackage{booktabs}       
\usepackage{amsfonts}       
\usepackage{nicefrac} 

\usepackage{microtype}      
\usepackage{lipsum}
\usepackage{graphicx}
\graphicspath{ {./images/} }

\usepackage{amssymb}
\usepackage{amsmath}

\usepackage{multirow}
 \usepackage{pifont} 
 \usepackage{tabularx}
\usepackage{array}
\usepackage{subcaption}
\usepackage{verbatim}
\usepackage{float}
 
\usepackage{threeparttable}
\usepackage{siunitx}
\usepackage[linesnumbered,ruled,vlined]{algorithm2e}
\usepackage[inkscapelatex=false]{svg}

\title{Generalizable Robustness Testing of DNN-Based Robotic Navigation Systems via XAI-Guided Search}

\author{
 Khizra Sohail \\
  Mondragon University\\
  Mondragon, Spain\\
  \texttt{ksohail@mondragon.edu} \\
   \And
 Miren Illaramendi \\
  Mondragon University\\
  Mondragon, Spain\\
  \texttt{millarramendi@mondragon.edu} \\
  \And
 Aitor Arrieta \\
Mondragon University\\
  Mondragon, Spain\\
  \texttt{aarrieta@mondragon.edu} \\
}

\begin{document}
\maketitle
\begin{abstract}
\textbf{Context:}
Deep Neural Networks (DNNs) are increasingly used to control Cyber-Physical Systems (CPSs), but small input perturbations may propagate into unsafe system-level behavior. Existing perturbation-generation approaches commonly optimize for individual images, potentially producing input-specific tests, and are often evaluated only in simulation without examining whether the identified failures transfer to physical systems.

\textbf{Objectives:}
This work aims to generate perturbation-based robustness tests that remain effective across different operational observations and to assess the extent to which their effects transfer from simulation to a physical robotic system.

\textbf{Methods:}
We propose an explainability-guided multi-objective evolutionary approach that generates sparse perturbations across a compact set of representative images. The images are selected by clustering visual and behavioral features, while Integrated Gradients are aggregated into a global attribution map that guides mutation toward influential image regions. We evaluate the approach through using a DNN-controlled LeoRover in Gazebo, conduct an ablation study of multi-image optimization and XAI guidance, and validate a stratified subset of perturbations on the real physical robot.

\textbf{Results:}
The proposed approach achieved a median success rate of 70.0\%, compared with 53.85\% for unguided search, and increased the median hypervolume from 0.65 to 0.73. Multi-image optimization increased the median success rate from 50.0\% to 57.5\%, while adding XAI guidance further increased it to 70.0\%, significantly outperforming both optimization variants. In the sim-to-real evaluation, simulation achieved a precision of 0.95 and a recall of 0.67, while simulated and physical failure times exhibited a significant positive correlation of 0.617.

\textbf{Conclusion:}
The results show that combining multi-image optimization with explainability-guided search improves the generation of effective robustness tests for DNN-based robotic systems. Simulation provides a useful mechanism for identifying and prioritizing transferable failures, but its limited recall demonstrates that physical validation remains necessary for safety-critical robustness assessment.
\end{abstract}





\section{Introduction}
\label{sec:Intro}

Cyber-Physical Systems (CPSs) integrate computational components with physical processes through sensing, computation, communication, and control. They are increasingly deployed in domains such as autonomous driving, mobile robotics, industrial automation, and other safety-critical applications. In these systems, Deep Neural Networks (DNNs) are progressively replacing or complementing conventional perception and control components because of their ability to learn complex mappings directly from high-dimensional sensory data~\cite{lecun2015deep, bojarski2016end}. For example, an autonomous navigation system may use a DNN to map camera images directly to steering commands. However, the behavior learned by these models is difficult to characterize exhaustively, and relatively small changes in their inputs may lead to unexpected variations in their outputs~\cite{szegedy2013intriguing}. When the output of the DNN directly influences a physical process, such variations may propagate to the system level and result in unsafe behaviors, such as lane departures or collisions. Consequently, systematically testing the robustness of DNN-based CPSs under perturbed operating conditions is essential before their deployment.

Several approaches have been proposed to generate perturbations that expose weaknesses in DNNs. These include gradient-based and black-box adversarial techniques~\cite{goodfellow2015explaining, carlini2017towards, alzantot2018genattack} and search-based techniques that explore the input space to identify failure-inducing conditions~\cite{harman2012search, tian2018deeptest}. Nevertheless, existing approaches commonly optimize a perturbation with respect to a single input image~\cite{su2019one, chen2019poba, williams2023black, tian2023imperceptible}. Although such a perturbation may effectively alter the output produced for that particular image, it may exploit image-specific characteristics and fail to generalize to other observations of the same operational scenario. This problem is particularly relevant for autonomous systems, where the controller receives a continuous sequence of images rather than a single isolated input.

Optimizing perturbations over multiple images can reduce this form of overfitting by requiring the generated perturbations to remain effective across different observations. However, directly evaluating every candidate perturbation over a large collection of images substantially increases the computational cost of the search. This limits the number of images that can be considered and, consequently, the diversity of operational situations represented during optimization. Moreover, mutation operators employed by evolutionary and black-box approaches are commonly unaware of which image regions influence the controller most strongly~\cite{alzantot2018genattack, chen2019poba, williams2023black}. As a result, a considerable part of the search effort may be spent modifying regions with little effect on the controller's behavior.

A further limitation concerns how the generated perturbations are evaluated. Most existing studies assess perturbations either using image-classification benchmarks or exclusively within a simulated environment~\cite{su2019one, chen2019poba, tian2023imperceptible}. Simulation provides a controllable and repeatable environment in which many test executions can be conducted at relatively low cost. Nevertheless, failures observed in simulation do not necessarily transfer directly to a physical system. Conversely, conducting the complete search on a physical robot would be costly, slow, difficult to reproduce, and potentially unsafe. Therefore, an important open problem is to determine whether perturbations generated and assessed in simulation preserve their effectiveness when transferred to the corresponding physical system.

To address these challenges, this paper presents an explainability-guided, multi-objective evolutionary framework for testing the robustness of DNN-based autonomous navigation systems. Instead of optimizing one perturbation independently for each image, the framework searches for a common perturbation structure that is evaluated over a representative set of images. The perturbation is composed of sparse circular regions and is optimized to simultaneously maximize its effect on the steering decisions of the controller and minimize its modification cost. Considering multiple images during the search encourages the resulting perturbations to capture weaknesses that are shared across different observations rather than characteristics that are specific to a single image. To improve the efficiency of this multi-image search, the framework incorporates information obtained through Integrated Gradients~\cite{sundararajan2017axiomatic}. Attribution maps are computed for the images considered during optimization and aggregated into a global map representing the image regions that most strongly influence the steering behavior of the controller. This information is then converted into a probabilistic mutation mechanism that prioritizes influential regions while preserving sufficient randomness to explore other parts of the input space. In this manner, explainability is not used only to interpret the decisions of an already trained model, but as operational information for directing the generation of robustness tests.

We evaluate the framework using a DNN-controlled LeoRover mobile robot and its corresponding Gazebo simulation environment. Through an ablation study, we separately investigate the effects of optimizing over multiple images and of guiding the evolutionary operators with attribution information. We then execute the generated perturbations repeatedly in simulation and validate a representative subset on the physical robot. This enables us to assess not only their capacity to induce system-level failures, but also the extent to which simulation-based observations transfer to the real system.

The main contributions of this paper are:

\begin{itemize}
\item A formulation of robustness testing that optimizes perturbations across representative operational scenarios rather than individual observations, reducing the image-specific overfitting of existing perturbation-generation approaches.

\item An explainability-guided search strategy that aggregates attribution information from multiple observations and exploits it to bias evolutionary search toward image regions that consistently influence the controller.

\item A search-based robustness testing framework that jointly optimizes perturbation effectiveness and perturbation cost, producing a diverse set of candidate robustness tests.

\item A comprehensive empirical evaluation, including an ablation study that isolates the contribution of multi-image optimization and XAI guidance, together with a sim-to-real validation on a physical robotic platform.

\end{itemize}

\section{Background}

In this section we explain basic concepts and background related to our study.

\subsection{DNN-Based Robotic Navigation Systems}

Robotic navigation systems operate as closed-loop CPSs in which sensory observations are continuously processed to determine control actions. In vision-based navigation, a camera captures an image $x_t$ of the environment at time $t$, and a controller $f$ maps this observation into a control command $u_t=f(x_t)$, such as the linear and angular velocity of the robot. The execution of $u_t$ changes the physical state of the robot and, consequently, the image received during the next control cycle. Therefore, errors in individual predictions may, or may not, propagate over time and ultimately lead to system-level failures.

At the model level, robustness can be assessed by measuring how input
variations affect the output of the DNN. However, a large change in the controller output does not necessarily imply that the robotic system will fail, while a sequence of comparatively small deviations may accumulate and cause unsafe behavior. Consequently, robustness testing of robotic navigation systems should complement model-level measurements with system-level execution outcomes.

\subsection{Search-based Software Testing}

Search-based software testing aims at recasting a software testing problem (e.g., test generation) into an optimization problem \cite{mcminn2011search,harman2015achievements}. For instance, in the context of test generation, instead of enumerating the complete input
space, a search algorithm iteratively generates and evaluates candidate test inputs according to one or more fitness functions. Each candidate represents a potential test case, while variation operators such as mutation are used to explore neighboring regions of the search space. The fitness functions guide the search toward inputs that are more likely to expose failures or exercise relevant system behavior.

Robustness-test generation commonly involves conflicting objectives. For example, a perturbation should have a substantial effect on the system while remaining small or localized. Multi-objective search addresses this by identifying a set of non-dominated solutions rather than aggregating all criteria into a single value. A solution dominates another when it is no worse in every objective and better in at least one, commonly known as Pareto-optimality. The resulting Pareto front represents alternative trade-offs from which testers can select according to their needs.

In DNN-based systems, search-based techniques have been used to explore high-dimensional input spaces and generate failure-inducing images~\cite{tian2018deeptest, alzantot2018genattack}. Candidate inputs can be modified through transformations or localized perturbations, while fitness functions quantify changes in model predictions, coverage, uncertainty, or system-level behavior. Nevertheless, the effectiveness of these approaches depends strongly on the representation, fitness functions, and information used to guide the search.

\section{XAI-guided search-based robustness testing}


\textcolor{black}{Our approach consists of three sequential phases. First, instead of generating perturbations from a single input image (as conducted in the literature~\cite{su2019one, liu2024beware}), we extract a compact set of representative images from the operational dataset, since single-image perturbation-based tests may overfit to one specific scene and may not reflect the diversity of situations encountered by the robot. Second, we compute a global attribution map over these representative images to identify the image regions that most consistently influence the DNN’s control decisions. Third, we use this attribution map to guide a multi-objective evolutionary search that generates universal perturbations, optimizing the trade-off between maximizing the deviation in the robot’s steering output and minimizing the number of perturbed pixels. This allows the approach to search for perturbation-based tests that are not only effective on isolated images, but that generalize across multiple representative navigation scenarios.}

\subsection{Phase 1: Data Clustering Pipeline}
\label{sec:phase1}
\textcolor{black}{During the first phase, we aim at selecting a subset of representative images to be used in the subsequent phases of the approach. This step is necessary because optimizing perturbation-based tests over the complete image dataset would be computationally expensive, while relying on a single image could lead to perturbations that are highly specific to one particular scene and therefore not representative of the robot’s broader operational conditions. To address this issue, we extract representative images through a clustering-based process that considers both visual information and behavioral context. The visual information captures the appearance of the scene, whereas the behavioral context captures the corresponding control behavior of the robot. By combining both sources of information, the selected images preserve diverse navigational situations while reducing redundant frames. The resulting representative image set provides a compact yet meaningful basis for generating perturbations that can generalize across multiple operational scenarios. Therefore, the objective of Phase I is not merely to reduce computational cost, but to derive a compact set of representative operational scenarios over which robustness tests are optimized. Optimizing over representative scenarios rather than individual observations encourages perturbations to capture weaknesses that persist across different manifestations of the same navigation behavior.}

    

    

\begin{figure}[ht]
    \centering
    
    \begin{subfigure}[b]{0.45\textwidth}
        \centering
        \includegraphics[width=\linewidth]{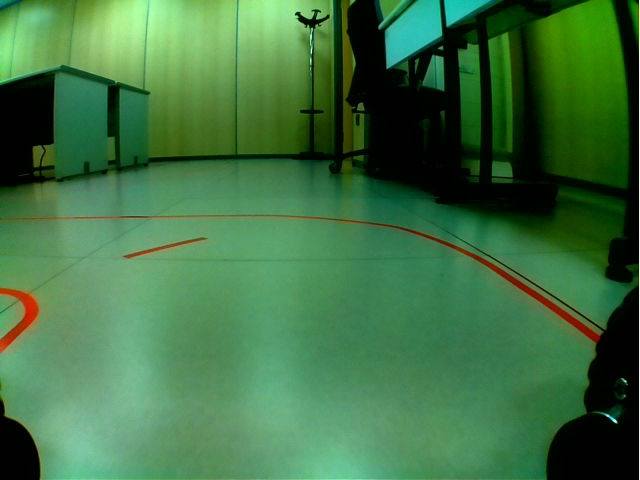}
        \caption{angular velocity = 0}
    \end{subfigure}
    \hfill
    \begin{subfigure}[b]{0.45\textwidth}
        \centering
        \includegraphics[width=\linewidth]{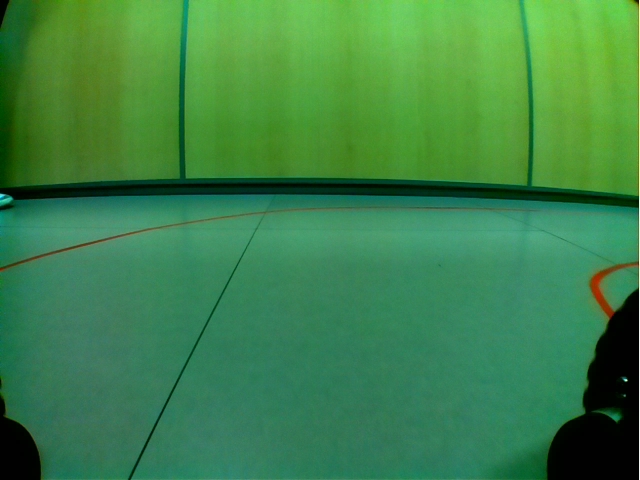}
        \caption{angular velocity = 0}
    \end{subfigure}

    \caption{Visually different navigational maneuvers with identical angular velocity}
    \label{fig:diff visual}
\end{figure}
\begin{figure}[ht]
    \centering
    
    \begin{subfigure}[b]{0.45\textwidth}
        \centering
        \includegraphics[width=\linewidth]{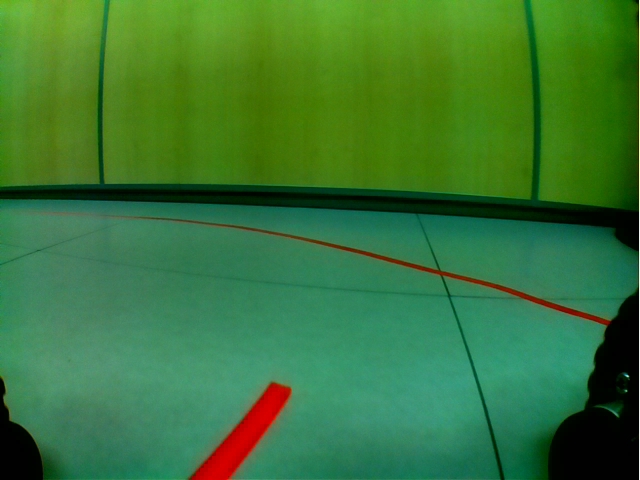}
        \caption{angular velocity = 0.61}
    \end{subfigure}
    \hfill
    \begin{subfigure}[b]{0.45\textwidth}
        \centering
        \includegraphics[width=\linewidth]{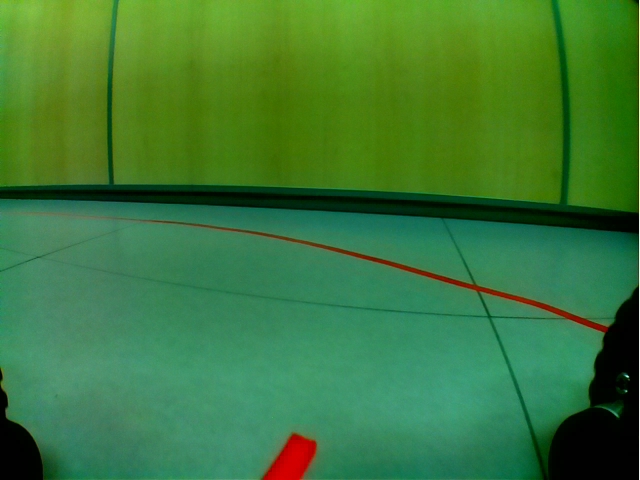}
        \caption{angular velocity = 1.09}
    \end{subfigure}

    \caption{Visually similar navigational maneuvers with different angular velocities}
    \label{fig:diff behavior}
\end{figure}

\textcolor{black}{The clustering pipeline combines visual features and behavioral context to group images that represent similar operational situations. For the visual component, we use DINOv2 (ViT-B/14) \cite{oquab2024dinov2} as a feature extractor, as it provides high-level semantic image embeddings without requiring task-specific labels. This is preferable to clustering raw pixels, which may be overly sensitive to low-level visual differences and fail to capture scene-level structure. The resulting visual embeddings are combined with behavioral information, such as the robot’s linear and angular velocities, to account for the fact that visual similarity and control behavior are not always aligned. For instance, two visually different scenes may have the same angular velocity if the robot has not yet started a maneuver (e.g., Figure~\ref{fig:diff visual}), while two visually similar scenes may require different angular velocities (e.g., Figure~\ref{fig:diff behavior}). Therefore, using either visual or behavioral information alone may produce misleading clusters. By constructing a joint feature space, the clustering process preserves both scene diversity and maneuver-related behavior, yielding a compact set of representative images for subsequent perturbation-based tests generation.}

\textcolor{black}{After constructing the joint visual-behavioral feature space, we apply dimensionality reduction before clustering. This step reduces the effect of the curse of dimensionality, under which distance-based measures become less discriminative in high-dimensional spaces~\cite{aggarwal2001surprising,beyer1999nearest}, while also lowering the computational cost of the subsequent clustering process. Following common practice in image clustering~\cite{allab2020considerably,wani2025review}, we retain the most informative components of the feature vectors and apply HDBSCAN to the reduced space. We use HDBSCAN because it does not require the number of clusters to be specified in advance, can identify clusters with different densities, and explicitly handles outliers by labeling them as noise~\cite{campello2013density,campello2015hierarchical}. Finally, we sample one image from each cluster. Since clustering has already grouped images according to their joint visual and behavioral similarity, this selection preserves coverage across the main operational situations while avoiding the redundancy of highly similar frames. The resulting reduced dataset makes the subsequent perturbation-based tests generation process computationally feasible without relying on a single, potentially unrepresentative image.}

\subsection{Phase II: Generation of Global Attribution Map}
\label{Phase2}

\textcolor{black}{The second phase generates a global attribution map that guides the subsequent evolutionary search. Rather than using explainability as a post-hoc interpretation tool, we exploit it as prior knowledge to guide the search process. Indeed, instead of selecting perturbation locations uniformly at random, we use feature attribution to identify image regions that are consistently influential for the DNN's control decisions. For an input image $x$ and model $f$, an attribution method $\Psi$ produces a relevance map $A_x = \Psi(f,x)$, where each pixel is assigned an importance score according to its contribution to the model output.}

\textcolor{black}{Although the framework is agnostic to the attribution method, we use Integrated Gradients (IG) because it provides dense pixel-level attributions and satisfies desirable axiomatic properties, such as sensitivity and implementation invariance~\cite{sundararajan2017axiomatic}. Dense attributions are particularly useful in our setting, where small visual elements such as lane markings, road edges, or turn indicators may strongly influence the robot's steering decisions. Following standard practice, we use a zero-valued image as the baseline input.}

\textcolor{black}{To obtain a global guidance signal, we aggregate the absolute IG attribution maps over the representative image set $D$ obtained in Phase I:
 \[
H_{ij} = \sum_{x \in D} |IG(x)_{ij}|.
\]}

\textcolor{black}{The resulting map $H$ highlights the spatial regions that most consistently influence the model across multiple operational scenarios, rather than only for a single image. This is important because the objective of our approach is to generate universal perturbations that generalize across representative navigation situations.
}

\textcolor{black}{Finally, we normalize $H$ and transform it into a probability distribution $W$:
\[
W_{ij} = \frac{H^{norm}_{ij}}{\sum_{p=1}^{H}\sum_{q=1}^{W} H^{norm}_{pq}}.
\]}

\textcolor{black}{This distribution is used in Phase III to guide mutation operators toward high-attribution regions. As a result, the search process remains stochastic, but it is biased toward image locations that are more likely to affect the DNN's steering output.}

\subsection{Phase III: XAI-Guided Multi-Objective Search Framework}
\label{Phase III}

\begin{algorithm}[ht]
\small
\setcounter{AlgoLine}{0}
\SetAlgoLined
\DontPrintSemicolon

\KwIn{Representative image dataset $\mathcal{D}_{rep}$, DNN under test $\mathrm{DNN}_{UT}$, importance distribution $W$, archive capacity $A_{\max}$, maximum number of iterations $T$}
\KwOut{Pareto archive $\mathcal{A}$}

\BlankLine
$\mathcal{A} \leftarrow \emptyset$\;

\For{$t = 1$ \KwTo $T$}{

    $\mathcal{S}_{parent} \leftarrow \textsc{Selection}(\mathcal{A})$\;

    $\mathcal{S}_{child} \leftarrow 
    \textsc{Mutation}(W, \mathcal{S}_{parent})$\;

    $\vec{F} \leftarrow 
    \textsc{Fitness Computation}(
    \mathcal{S}_{child}, W,
    \mathcal{D}_{rep},
    \mathrm{DNN}_{UT})$\;

    $\mathcal{A} \leftarrow 
    \textsc{UpdateArchive}(\vec{F}, \mathcal{S}_{child}, A_{\max})$\;
}

\Return{$\mathcal{A}$}\;

\caption{XAI-Guided Archive-Based Multi-Objective Search}
\label{alg:MOEA}
\end{algorithm}

\textcolor{black}{The third phase generates localized perturbations through a multi-objective search process. Given the representative image set obtained in Phase~I and the attribution-based importance distribution generated in Phase~II, the goal is to identify perturbations that induce strong deviations in the DNN's output while modifying as few pixels as possible. This naturally leads to a multi-objective formulation, where perturbation effectiveness and cost are optimized simultaneously. Each candidate solution encodes a set of localized image regions in which noise is injected, and the same perturbation is applied across all representative images to promote generalization beyond a single scene. Rather than adopting a full population-based MOEA such as NSGA-II, we use a lightweight archive-based search strategy. This choice is motivated by the high cost of evaluating each candidate perturbation across multiple representative images and by the variable-length structure of our solutions, where perturbation regions can be added, removed, moved, or resized. The archive allows the search to retain a set of non-dominated perturbations while immediately discarding dominated candidates. Thus, our approach preserves the main benefit of Pareto-based evolutionary optimization (i.e., maintaining diverse trade-offs between DNN output deviation and perturbation cost) while avoiding the overhead of maintaining and sorting a full population at each generation.}

\textcolor{black}{Algorithm \ref{alg:MOEA} summarizes the proposed search process. The algorithm receives as input: (i) the representative input dataset $\mathcal{D}_{rep}$ obtained in Phase~I; (ii) the DNN under test $DNNUT$; (iii) the importance distribution $W$ computed in Phase~II, which guides the spatial placement of perturbations; (iv) the maximum archive capacity $A_{\max}$; and (v) the maximum number of search iterations $T$. The output is a Pareto archive containing non-dominated perturbations that represent different trade-offs between the DNN output and perturbation cost. The algorithm begins by initializing the archive (Line~1), followed by entering a loop that guides the iterative generation, evaluation, and selection of candidate perturbations over a fixed number of search iterations (Lines~2-7).}

\textcolor{black}{Within the loop, the first step is the selection of a parent solution from the archive (Line~3). Since the archive only stores non-dominated solutions, each archived candidate represents a potentially useful trade-off between perturbation-based test effectiveness and perturbation cost. We therefore sample the parent uniformly at random from the archive, rather than prioritizing a specific objective or region of the Pareto front. This prevents the search from prematurely focusing on, for example, highly aggressive perturbations or very small perturbations only, and allows different parts of the trade-off space to be explored. }

\textcolor{black}{After the parent is selected, the mutation process is performed. This process takes as input the parent solution $\mathcal{S}_{parent}$ and the importance distribution $W$ obtained in Phase~II, and produces a new candidate solution $\mathcal{S}_{child}$. The mutation process is responsible for modifying the structure of the perturbation, either by adding new perturbed regions, removing existing ones, moving them to different image locations, or changing their size. In our algorithm, one mutation operator is selected at random from the set $\{\emph{Add}, \emph{Remove}, \emph{Move}, \emph{Scale}\}$. \textcolor{black}{When the image has no perturbation, the only possible mutation operator is \emph{Add}, since the remaining operators require an existing perturbation to modify. The \textit{Add} operator initializes the first perturbation region. After each mutation, Gaussian noise sampled from a zero-mean normal distribution $\mathcal{N}(0,\sigma^{2})$ is injected into the pixels belonging to the current perturbation regions before the candidate solution is evaluated. Consequently, \textit{Add} creates the initial perturbation, while the remaining operators modify its spatial properties.} Figure \ref{fig:Illustration Mutation} provides illustrative image examples of each of these perturbations. 
\begin{figure}
    \centering
    \includegraphics[width=0.95\linewidth]{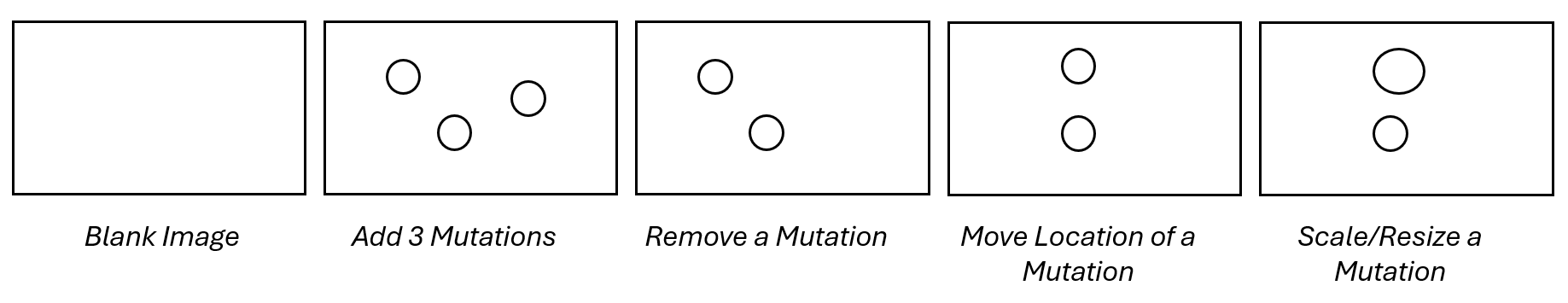}
    \caption{Illustration of the application of mutation operators}
    \label{fig:Illustration Mutation}
\end{figure}
This random operator selection allows the search to alternate between expanding, simplifying, relocating, and refining candidate perturbations. The role of $W$ is to guide the spatial decisions made by these mutation operators. In particular, when new regions are added or existing regions are moved, their locations are sampled according to the probability distribution $W$, making high-attribution regions more likely to be selected. Similarly, when regions are removed or resized, the attribution values are used to favor the removal or reduction of regions located in less influential areas. Therefore, the mutation process remains stochastic, but it is no longer purely random: it is biased toward image regions that are more likely to affect the DNN's output. This design enables the search to exploit the model-specific information obtained through XAI while still preserving exploration of different perturbation structures.}

\textcolor{black}{Once $\mathcal{S}_{child}$ has been generated, it is evaluated over the entire representative input dataset $\mathcal{D}_{rep}$ (Line 5). To this end, the perturbation encoded by $\mathcal{S}_{child}$ is applied to each image $x \in \mathcal{D}_{rep}$, producing a perturbed image $\tilde{x}$. Both the original and perturbed images are then provided to the DNN under test, denoted as $DNN_{UT}$, in order to measure the effect of the perturbation on the model's control output. The quality of $\mathcal{S}_{child}$ is assessed using two fitness objectives: effectiveness and cost. The effectiveness objective $f_1$ measures the average deviation induced in the DNN's output across the representative dataset:
\[
f_1(\mathcal{S}_{child}) =
\frac{1}{|\mathcal{D}_{rep}|}
\sum_{x \in \mathcal{D}_{rep}}
\left|
DNN_{UT}(\tilde{x})_{\omega} - DNN_{UT}(x)_{\omega}
\right|,
\]
where $DNN_{UT}(x)_{\omega}$ denotes the output predicted by the model for image $x$. This objective is maximized, since larger deviations indicate stronger perturbation effects. The cost objective $f_2$ measures the number of pixels modified by the perturbation:
\[
f_2(\mathcal{S}_{child}) =
\left|\{p \mid \delta(p) \neq 0\}\right|.
\]
This objective is minimized, since lower values correspond to more localized and less intrusive perturbations. Therefore, each candidate solution is evaluated according to the trade-off between maximizing DNN output and minimizing the number of perturbed pixels.}

\textcolor{black}{The last step of each iteration is the archive update (Line 6). This step takes as input the candidate solution $\mathcal{S}_{child}$, its fitness vector $\vec{F}(\mathcal{S}_{child})$, the current archive $\mathcal{A}$, and the maximum archive capacity $A_{\max}$. The archive is updated according to Pareto dominance. If $\mathcal{S}_{child}$ is dominated by any solution already stored in $\mathcal{A}$, it is discarded, since it does not improve the current approximation of the Pareto front. Otherwise, $\mathcal{S}_{child}$ is inserted into the archive, and any existing archive solutions dominated by $\mathcal{S}_{child}$ are removed. In this way, the archive incrementally maintains the set of non-dominated perturbations found during the search. Since the number of non-dominated solutions may grow beyond the allowed archive capacity, we apply a diversity-preservation mechanism whenever $|\mathcal{A}| > A_{\max}$. Specifically, crowding distance is used to estimate how densely populated each region of the objective space is. The solution located in the most crowded region is removed, preserving a diverse approximation of the Pareto front. This prevents the archive from concentrating only on a narrow subset of solutions and maintains different trade-offs between perturbation-based test effectiveness and perturbation cost.}

\textcolor{black}{The search process continues until the maximum number of iterations $T$ is reached. At that point, no further candidate perturbations are generated, and the archive $\mathcal{A}$ is returned as the output of the algorithm. Since the archive has been updated throughout the search using Pareto dominance and crowding-distance pruning, it contains a diverse set of non-dominated perturbations that represent different trade-offs between perturbation-based test effectiveness and perturbation cost.}

\section{Experimental Setup}

This section explains the experimental setup of our approach when applying it into an autonomous robotic platform. Specifically, we aim to answer the following three research questions (RQs): 

\begin{itemize}

\item \textbf{RQ1 -- Sanity Check:} 
\textit{Does XAI-guided evolutionary search outperform unguided search in generating high-impact perturbations?} This RQ serves as a sanity check for the overall search strategy. We compare the proposed approach against an unguided baseline that uses the same perturbation representation, objectives, and search budget, but does not use attribution-guided mutation or an online Pareto archive to retain promising solutions during the search. The goal is to verify whether combining XAI-guided mutation with archive-based preservation of non-dominated solutions provides a measurable advantage over uninformed exploration.

\item \textbf{RQ2 -- Ablation of the Optimization Formulation}
\textit{How does the optimization formulation affect the reliability and quality of generated perturbations across operational scenarios?} This RQ analyzes the contribution of the main formulation choices in our approach. In particular, we compare single-image optimization, multi-image optimization, and XAI-guided multi-image optimization. Single-image optimization may produce perturbations that are effective for individual frames but overfit to their specific visual characteristics. Multi-image optimization addresses this limitation by requiring each candidate perturbation to be evaluated across the representative image set, thereby favoring perturbations that generalize across scenarios. Finally, XAI-guided multi-image optimization evaluates whether attribution-guided mutation further improves the quality of the generated perturbations. This RQ therefore assesses how each formulation affects perturbation-based test reliability, success rate, and the trade-off between perturbation cost and perturbation-based test severity.

\item \textbf{RQ3 -- Sim-to-Real Transfer:}
\textit{To what extent do perturbations generated with our approach and evaluated in simulation transfer to physical robot execution?} This RQ evaluates whether simulation-based outcomes are predictive of physical robot behavior. Since executing all perturbations on the real robot is costly and time-consuming, simulation is used as the main evaluation environment. However, the sim-to-real gap may cause perturbations that appear effective in simulation to behave differently on hardware. We therefore assess whether simulated perturbation outcomes are consistent with physical execution, considering both successful perturbations that induce out-of-bounds behavior and unsuccessful perturbations that remain benign.
\end{itemize}


\subsection{Case Study System}

\textcolor{black}{We evaluate our approach using a LeoRover autonomous mobile robot operating on a closed-loop track delimited by colored tape. The robot is controlled by an image-based DNN that receives a front-facing camera image as input and outputs the corresponding control commands, namely linear and angular velocity. These commands are then sent to the rover through a ROS-based control pipeline. This setting provides a controlled yet physically executable case study for evaluating perturbation-based robustness testing of DNN-driven robotic navigation.}

\begin{figure}[ h ]
    \centering
    \includegraphics[width=0.63\linewidth, height= 6.8cm]{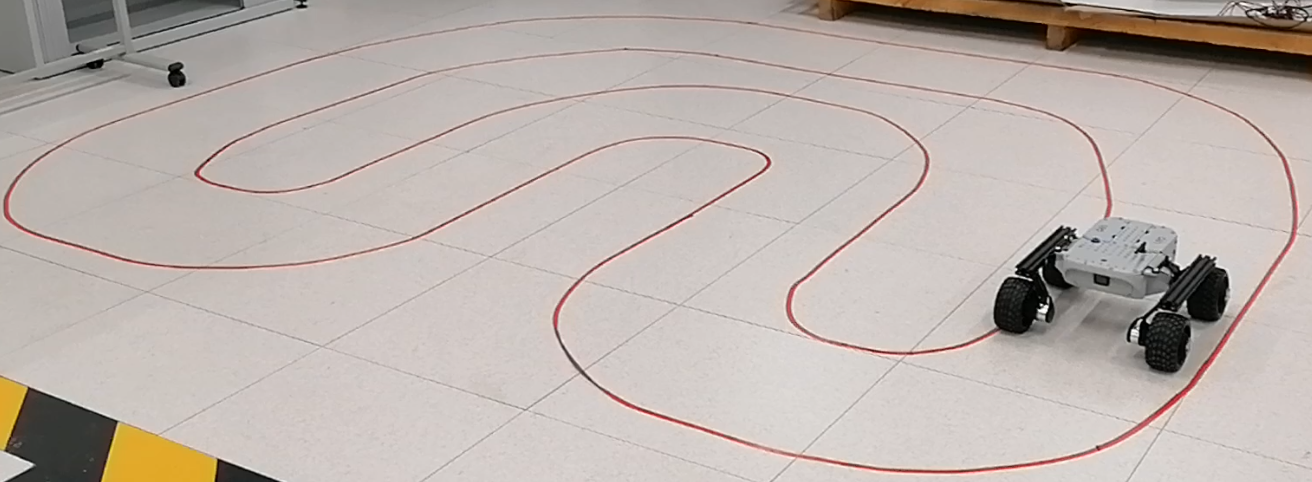}
    \caption{LeoRover operating on the physical circuit used in our experiments}
    \label{fig:leorover circuit}
\end{figure}

\textcolor{black}{The DNN controller is trained to follow the track from camera images. In our experiments, we use a model trained for 100 epochs using both the dataset provided by the LeoRover vendor~\cite{leorover_2023} and the dataset introduced in~\cite{arrieta2023deep}, resulting in a training set of 16,048 images. The model architecture consists of five convolutional layers followed by two dense layers. Figure~\ref{fig:leorover circuit} shows the physical circuit used in our evaluation.
}

\begin{figure}[h]
    \centering
    \includegraphics[width=0.65\linewidth, height=7cm, keepaspectratio]{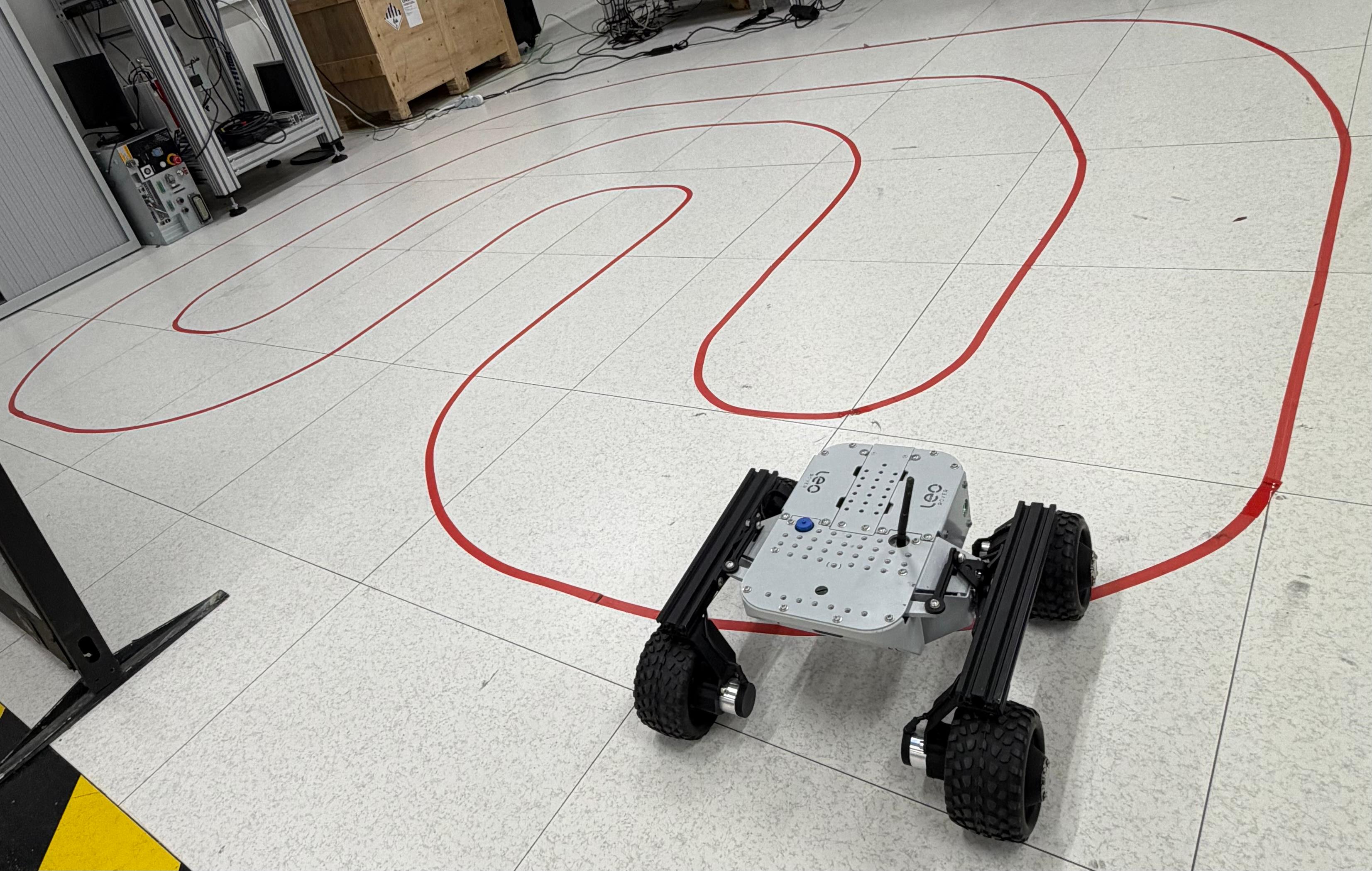}
    \caption{Example of an out-of-bounds failure in the physical LeoRover setup}
    \label{fig:OOB Leorover}
\end{figure}

We consider an execution to fail when the rover goes out of bounds. For the physical robot, we follow the definition used in MarMot~\cite{ayerdi2024marmot}: the rover is considered out of bounds when both front wheels cross the circuit lines and the line is no longer visible from the camera view. This criterion captures failures that are difficult for the controller to recover from, since the visual cue required for lane following is no longer available. Figure~\ref{fig:OOB Leorover} illustrates this failure condition.

\begin{figure}[h]
    \centering
    \includegraphics[width=0.67\linewidth, height = 7cm , keepaspectratio]{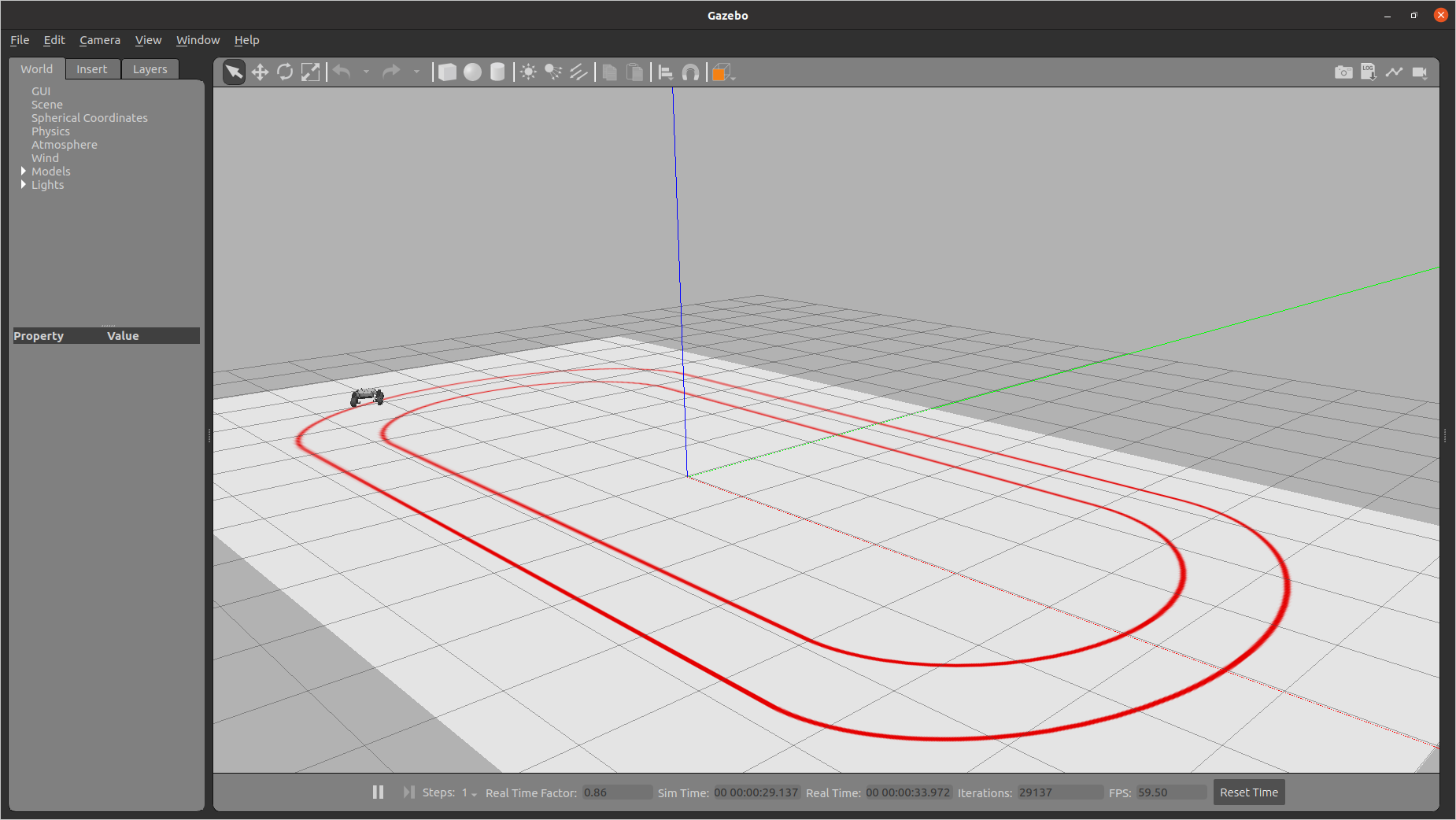}
    \caption{Example of an out-of-bounds failure in the Gazebo simulation environment}
    \label{Simulator}
\end{figure}

In addition to physical executions, we use the LeoRover Gazebo simulator to evaluate perturbations under safe and repeatable conditions. Simulation allows us to execute a large number of perturbation-based tests before selecting a subset for physical validation. Figure~\ref{Simulator} shows an example of an out-of-bounds failure in the simulated environment.

In simulation, the out-of-bounds condition is detected geometrically by measuring the lateral distance between the rover center point and the road centerline. An out-of-bounds event is raised when this distance exceeds the maximum allowable offset, following the criterion used in~\cite{ayerdi2024marmot}.

\subsection{Dataset and Experimental Parameters}

\textcolor{black}{We evaluate our framework using the LeoRover driving dataset introduced in~\cite{arrieta2023deep}. The dataset contains 16,048 images captured from a forward-facing camera during autonomous navigation, together with the corresponding control outputs produced by the DNN controller. The dataset covers different navigation situations, including straight segments, curves, and varying environmental conditions.}

\textcolor{black}{As described in Section~\ref{sec:phase1}, applying the search process directly to the full dataset would be computationally expensive. Therefore, we use the clustering pipeline introduced in Phase~I to obtain a representative dataset of 51 images, selecting one image from each cluster. The clustering was performed using PCA to reduce the joint visual--behavioral feature vectors to 50 components, followed by HDBSCAN with a minimum cluster size of 10. This reduced dataset is used throughout the search process to make perturbation generation computationally feasible while preserving diversity across operational scenarios.}

\textcolor{black}{Table~\ref{tab:experimental_parameters} summarizes the main experimental parameters. Unless otherwise stated, the same configuration is used across all variants and baselines to ensure a fair comparison. Each search execution runs for 1,000 iterations, and each configuration is repeated for 30 independent runs, following common recommendations for randomized algorithms~\cite{arcuri2011practical}. Candidate perturbations are constrained to contain at most 10 localized circular regions, each with a radius sampled between 3 and 10 pixels. These bounds limit the perturbation size so that less than 10\% of the image pixels are modified. The Gaussian noise scale is set to $\sigma=0.1$, following prior work~\cite{yoon2023learning}, and the archive size is bounded to 20 solutions for archive-based variants.}

\begin{table}[h]
\centering
\caption{Experimental parameters used in the evaluation.}
\label{tab:experimental_parameters}
\begin{tabularx}{\columnwidth}{>{\raggedright\arraybackslash}X l}
\toprule
\textbf{Parameter} & \textbf{Value} \\
\midrule
Original dataset size & 16,048 images \\
Representative dataset size & 51 images \\
Dimensionality reduction & PCA, 50 components \\
Clustering algorithm & HDBSCAN, minimum cluster size $=10$ \\
Iterations per run & 1,000 \\
Independent runs & 30 \\
Maximum perturbation regions & 10 \\
Region radius & 3--10 pixels \\
Maximum perturbed pixels & $<10\%$ of the image \\
Noise scale $\sigma$ & 0.1 \\
Archive capacity $A_{\max}$ & 20 solutions \\
IG Interpolation steps & 50 \\
\bottomrule
\end{tabularx}
\end{table}

\begin{figure}[ht!]
    \centering
    \includegraphics[width=1\linewidth]{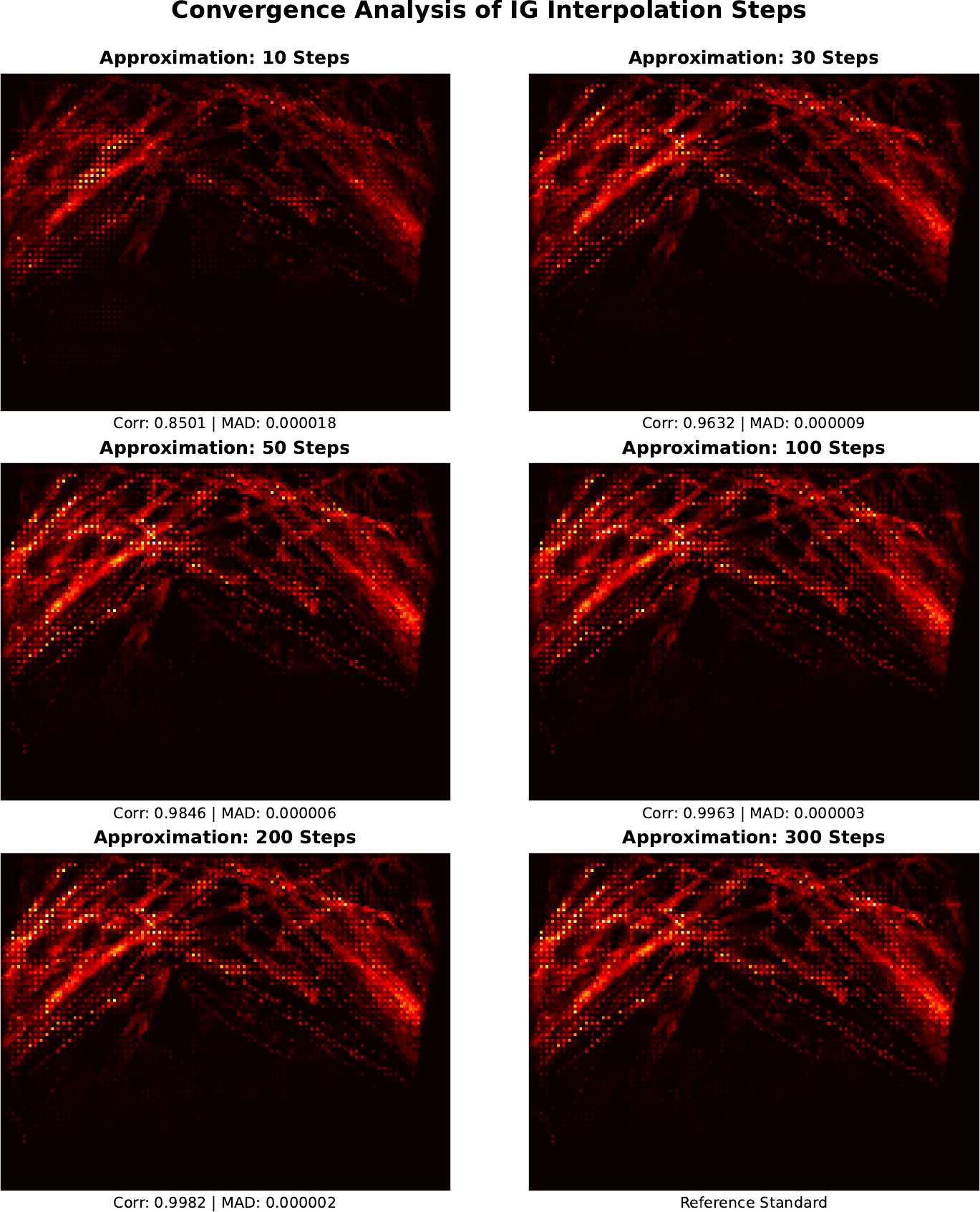}
    \caption{Convergence analysis of Integrated Gradients attribution maps for different numbers of interpolation steps.}
    \label{fig:ig_convergence}
\end{figure}

\textcolor{black}{For the proposed XAI-guided variant, we used 50 interpolation steps to compute Integrated Gradients. This value was selected empirically by comparing attribution maps obtained with different numbers of steps against a reference approximation computed with $m=300$. As shown in Figure~\ref{fig:ig_convergence}, the attribution maps stabilize rapidly: using $m=50$ yields a Pearson correlation of 0.9846 with the reference map and a mean absolute difference of 0.000006. This provides a good trade-off between attribution quality and computational cost.}

\subsection{Unguided Search Baseline (RQ1)}
\label{sec:unguided search baseline}
\textcolor{black}{To assess whether the proposed archive-based XAI-guided search provides a meaningful advantage over uninformed exploration, we compare it against an unguided search baseline. The baseline uses the same perturbation representation, mutation operators, objectives, search budget, and evaluation pipeline as the proposed approach. However, it does not use the attribution distribution $W$ to guide spatial decisions, nor does it maintain an online Pareto archive during the search. Instead, perturbation regions are sampled uniformly at random over the image space, and all generated candidates are stored during execution. After the search budget is exhausted, Pareto-optimal solutions are extracted post hoc using the same objectives as in the proposed approach: maximizing the average deviation in the DNN output across the representative image set and minimizing the number of perturbed pixels. This baseline therefore represents an uninformed perturbation generation strategy, allowing us to evaluate whether XAI-guided mutation and archive-based preservation of non-dominated solutions improve the discovery of effective perturbation-based robustness tests.}

\subsection{Ablation Study (RQ2)}

\textcolor{black}{To evaluate the contribution of the main design choices in our approach, we compare the proposed method against two ablated variants and one unguided baseline. The goal is to assess the impact of: (i) optimizing perturbations for a single image versus multiple representative images, (ii) using an archive-based evolutionary search, and (iii) guiding mutation with the attribution distribution obtained in Phase~II.}

\begin{itemize}

\item \textbf{Variant I: Single-Image Optimization.}
\textcolor{black}{This variant optimizes perturbations independently for each representative image. It uses the same perturbation representation and objective functions as the proposed approach, but each candidate is evaluated only on one image. This variant allows us to assess whether perturbations optimized for individual inputs generalize to broader operational scenarios, or whether they overfit to the visual characteristics of a single frame.} 

\item \textbf{Variant II: Multi-Image Optimization.}
\textcolor{black}{This variant evaluates each candidate perturbation across the full representative image dataset. Therefore, the same perturbation must induce deviations across multiple operational scenarios. Unlike the proposed approach, spatial mutation decisions are sampled uniformly at random rather than guided by the attribution distribution. This variant isolates the effect of moving from single-image to multi-image optimization.}

\textcolor{black}{
Together, these variants allow us to quantify the individual and combined contributions of search strategy, generalization, and explainability.
\begin{table}[h]
\centering
\caption{Ablation study configurations. \checkmark denotes the component is present.}
\resizebox{\columnwidth}{!}{%
\begin{tabular}{lccc}
\toprule
\textbf{Configuration} & \textbf{Multi-Image} & \textbf{Evolutionary Optimization} & \textbf{XAI Guidance} \\
\midrule
Unguided Search (Baseline)     & \checkmark & $\times$ &
$\times$ \\
Variant I (Single-Image MOEA)  & $\times$ & \checkmark & 
$\times$ \\
Variant II (Multi-Image MOEA)  & \checkmark & \checkmark & 
$\times$ \\
Proposed Approach              & \checkmark & \checkmark & \checkmark \\
\bottomrule
\end{tabular}%
}
\end{table}}

\end{itemize}
\subsection{Sim-to-Real Transfer of the Generated Perturbations (RQ3)}
\label{sec:Physical Robot Validation}
\textcolor{black}{All configurations were evaluated in the Gazebo simulator. In addition, we validated a subset of perturbations generated by the proposed approach on the physical LeoRover robot. This physical validation was limited to the proposed approach because hardware execution is substantially more time-consuming than simulation: each run requires manual setup, battery monitoring, robot reset, and verification under nominal conditions. Therefore, the goal of this experiment is not to compare all configurations on hardware, but to assess whether perturbation-induced failures observed in simulation can also be reproduced on the physical robot.}

\textcolor{black}{From the 600 perturbations generated by the proposed approach and evaluated in simulation, we selected 60 perturbations for physical validation. The selection was stratified according to the simulation outcome. For perturbations that caused an out-of-bounds event in simulation, we selected 10 with the shortest failure time, 10 with intermediate failure time, and 10 with the longest failure time. For perturbations that did not cause an out-of-bounds event, we selected 10 with low perturbation cost, 10 with intermediate perturbation cost, and 10 with high perturbation cost. This sampling strategy covers both successful and unsuccessful simulation outcomes, as well as different levels of severity and perturbation cost.}

\textcolor{black}{Before each physical execution, the robot was checked to ensure stable operating conditions. Battery voltage was monitored throughout the experiment, and the robot was recharged }\textcolor{black}{ whenever voltage dropped below 12V. The nominal range of LeoRover battery is defined to be 10.8 V~\cite{fictionlab_leo_specification}}.
\textcolor{black}{ After each recharge or system reconnection, the rover was required to complete two nominal laps without perturbations before continuing the validation. These checks reduce the risk of attributing failures to confounding factors such as low battery, system instability, or residual effects from previous executions.}

\textcolor{black}{Each selected perturbation was executed once on the physical robot. Table~\ref{tab:physical_execution_summary_aitor} summarizes the selection strategy.}

\begin{table}[h]
\centering
\caption{Selection of perturbations for physical robot validation.}
\label{tab:physical_execution_summary_aitor}
\begin{tabular}{llc}
\toprule
\textbf{Simulation outcome} & \textbf{Selection criterion} & \textbf{Number} \\
\midrule
OOB & Short failure time & 10 \\
OOB & Intermediate failure time & 10 \\
OOB & Long failure time & 10 \\
No OOB & Low perturbation cost & 10 \\
No OOB & Intermediate perturbation cost & 10 \\
No OOB & High perturbation cost & 10 \\
\midrule
\multicolumn{2}{r}{\textbf{Total}} & \textbf{60} \\
\bottomrule
\end{tabular}
\end{table}

\subsection{Evaluation Metrics}
\label{sec:evaluation_metrics}
\textcolor{black}{The output of each search execution is a Pareto archive $\mathcal{A}$ containing candidate perturbations. To evaluate these perturbations at the system level, we executed each perturbation $\delta \in \mathcal{A}$ in the Gazebo simulator and record whether the rover goes out of bounds. \textcolor{black}{Because simulator execution is stochastic, each archive was replayed twice before aggregating the evaluation metrics.} This post-search simulation step is important because the optimization process itself uses DNN-output deviation as a proxy objective, whereas the final evaluation focuses on whether the generated perturbations induce observable robotic failures. For each perturbation $\delta$, the simulator returns a binary outcome:}
\[
OOB(\delta) =
\begin{cases}
1, & \text{if the rover goes out of bounds},\\
0, & \text{otherwise}.
\end{cases}
\]
\textcolor{black}{Using this outcome, we compute the \textit{Success Rate} of an archive as the proportion of archived perturbations that induce an out-of-bounds event:}
\[
SR(\mathcal{A}) =
\frac{|\{\delta \in \mathcal{A} \mid OOB(\delta)=1\}|}{|\mathcal{A}|}
\times 100.
\]
\textcolor{black}{where higher success rates indicate that a larger fraction of the generated perturbations lead to system-level failures.}

\textcolor{black}{For each successful perturbation, we also record the \textit{failure time}, denoted as $\tau(\delta)$, which measures the time elapsed from the beginning of the simulation until the rover goes out of bounds. Lower failure times indicate more effective perturbations, since the failure occurs earlier during execution. Failure time is only defined for successful perturbations; perturbations that do not induce an out-of-bounds event are considered unsuccessful and are accounted for through the success rate.}

\textcolor{black}{In addition, we compute the \textit{perturbation cost}, denoted as $c(\delta)$, as the number of pixels modified by the perturbation:
\[
c(\delta) = |\{p \mid \delta(p) \neq 0\}|.
\]
where lower values are prefered as they correspond to less intrusive perturbations.}

\textcolor{black}{Finally, for each set of successful solutions in the archive we derive a Pareto front computed over two objectives: minimizing failure time $\tau(\delta)$ and minimizing perturbation cost $c(\delta)$. A perturbation is therefore preferred if it induces an out-of-bounds event faster and with fewer modified pixels. This Pareto front captures the trade-off between failure severity and perturbation cost among the successful perturbation-based robustness tests. When needed for aggregate comparison, we summarize the quality of this front using hypervolume after normalizing both objectives, with higher hypervolume indicating a better trade-off between low failure time and low perturbation cost.}
Both objectives are minimized (lower failure time and lower pixel changes are better). To enable fair comparison across different scales, each objective was normalized to $[0,1]$ using the global maximum observed across all algorithms and simulation runs:
\begin{align}
\tau_{\text{norm}} &= \frac{\tau}{\max_{\tau}} && \text{($\max_{\tau}$: max failure time)} \\
\text{pixel}_{\text{norm}} &= \frac{\text{pixel\_changes}}{\max_{\text{pixel}}} && \text{($\max_{\text{pixel}}$: max pixel changes)}
\end{align}
This normalization maps the worst observed value to $1$ and the best to $0$. Hypervolume was then calculated with the reference point $(1,1)$ in this normalized space, representing the worst possible combination of both objectives. Since both objectives are minimized, the hypervolume indicator measures the area dominated by the Pareto front relative to this reference point, with higher values (closer to $1$) indicating better overall performance. This approach follows standard practices in multi-objective optimization~\cite{zitzler2004indicator, cao2017uncertainty}.

\textcolor{black}{For the physical validation experiment in RQ3, the analysis was different because only a stratified subset of perturbations generated by the proposed approach was executed on the physical robot. We therefore assessed the sim-to-real consistency by comparing the simulator outcome and the physical outcome for each selected perturbation. We reported the resulting confusion matrix and derive standard classification metrics, including precision, recall, false positive rate, and F1-score.}
\textcolor{black}{
Let $Y_{\text{sim}} \in \{0, 1\}$ denote the simulator prediction and $Y_{\text{real}} \in \{0, 1\}$ denote the physical robot outcome, where $0$ indicates safe lap completion (no out-of-bounds) and $1$ indicates a navigation failure (out-of-bounds). The confusion matrix comprises four mutually exclusive categories:
\begin{itemize}
    \item \textbf{True Negative (TN):} Both the simulator and the physical robot agree on a safe outcome ($Y_{\text{sim}} = 0$, $Y_{\text{real}} = 0$). This category represents perturbations that are correctly identified as benign by the simulation pipeline.
    \item \textbf{False Positive (FP):} The simulator predicts a failure, but the physical robot completes the lap safely ($Y_{\text{sim}} = 1$, $Y_{\text{real}} = 0$). This category corresponds to conservative errors, where simulation raises an alarm that does not materialize on hardware.
    \item \textbf{False Negative (FN):} The simulator predicts a safe outcome, but the physical robot experiences a failure ($Y_{\text{sim}} = 0$, $Y_{\text{real}} = 1$). This category constitutes the most critical error type, as it indicates real-world vulnerabilities that evade detection during simulation-based screening.
    \item \textbf{True Positive (TP):} Both the simulator and the physical robot agree on a failure ($Y_{\text{sim}} = 1$, $Y_{\text{real}} = 1$). This category represents perturbations whose effect successfully transfers from simulation to the physical platform.
\end{itemize}
}

\subsection{Statistical Analysis} 

Each optimization configuration was executed for 30 independent evolutionary runs to account for the stochastic nature of evolutionary search, as recommended by Arcuri and Briand~\cite{arcuri2011practical}. Each run consisted of 1000 iterations and produced a final Pareto archive containing non-dominated perturbation structures. These archives represent the output of the search process and were subsequently evaluated at the system level using the Gazebo simulator. Since simulator execution contains stochastic variations, each generated Pareto archive was executed twice in simulation. Therefore, each configuration resulted in 60 simulator evaluations (30 evolutionary runs × 2 simulator repetitions).

\textcolor{black}{As described in Section~\ref{sec:evaluation_metrics}, each archive is evaluated in simulation to compute two aggregate metrics: Success Rate and Hypervolume. Success Rate measures the proportion of archived perturbations that induce an out-of-bounds event, whereas Hypervolume summarizes the quality of the Pareto front obtained from the successful perturbations in terms of failure time and perturbation cost. Each perturbation is simulated twice, as the simulator is stochastic. Since these metrics are computed over stochastic search executions, we first assessed their normality using the Shapiro-Wilk test. Several metric distributions violated the normality assumption; therefore, we used non-parametric statistical tests for the comparison of configurations. We first applied the Kruskal-Wallis test to determine whether statistically significant differences existed among the evaluated configurations. When significant differences were observed, we performed pairwise comparisons using the Mann-Whitney U test. To control the family-wise error rate introduced by multiple pairwise comparisons, we applied a Bonferroni correction with $\alpha=0.05$. Since we compare four configurations, this results in six pairwise comparisons per metric and an adjusted significance threshold of $\alpha_{adj}=0.0083$.}

\textcolor{black}{In addition to $p$-values, we report the Vargha--Delaney effect size $\hat{A}_{12}$ to quantify the magnitude of the observed differences. For a comparison ``A vs. B'', $\hat{A}_{12} > 0.5$ indicates that A tends to obtain higher values than B, whereas $\hat{A}_{12} < 0.5$ indicates that B tends to obtain higher values than A. For both Success Rate and Hypervolume, higher values are better. }

\subsection{Computational Resources For Experiment}
To support the computationally intensive requirements of the evolutionary algorithm and the associated TFLite model inference, all experiments were conducted on a personal laptop system. The system is equipped with an Intel(R) Core(TM) Ultra 7 155H processor operating at 1.40 GHz and 32.0 GB of RAM, which was sufficient to handle the iterative computational workload effectively. All experiments were conducted on the CPU within a 64-bit x64-based operating system environment to ensure stable and reliable performance throughout the experimental phase.
\section{Analysis of the Results and Discussion}
We now analyze the results and discuss the main findings of the experiment.
\subsection{RQ1 -- Sanity Check}
\begin{figure}[ht]
    \centering    
    \includegraphics[width=0.64\linewidth]{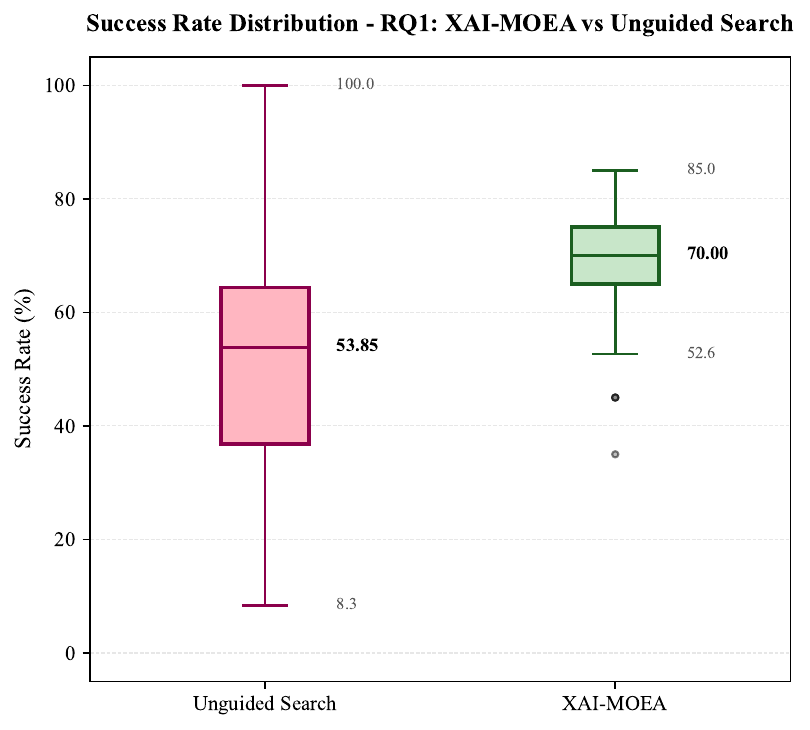}
    \caption{Success Rate comparison between XAI-MOEA and Unguided Search over 30 independent runs.}
    \label{fig:rq1_sr}
\end{figure}

\textcolor{black}{In RQ1 we conducted a sanity check by comparing our approach with an unguided baseline. The proposed approach clearly improved the failure-inducing capability of the generated perturbations, as can be seen in Figure \ref{fig:rq1_sr}. Our approach achieved a median success rate of 70.0\%, compared with 53.85\% for Unguided Search. Moreover, the distribution of results was more stable for our approach: its lower whisker remained at 52.6\%, whereas Unguided Search dropped to 8.3\% in the worst-performing runs. This indicated that the proposed approach not only generated more failure-inducing perturbations on average, but also reduced the risk of obtaining ineffective perturbation archives in individual runs. The difference was statistically significant according to the Mann--Whitney U test after Bonferroni correction ($p < 0.00001$). The Vargha-Delaney effect size for the comparison Unguided Search vs. our approach was $\hat{A}_{12}=0.2222$ for the success rate metric, which, given the comparison direction, indicated that XAI-MOEA obtained higher success rates in the majority of pairwise comparisons.}
\begin{figure} [ht]
    \centering    
    \includegraphics[width=0.65\linewidth]{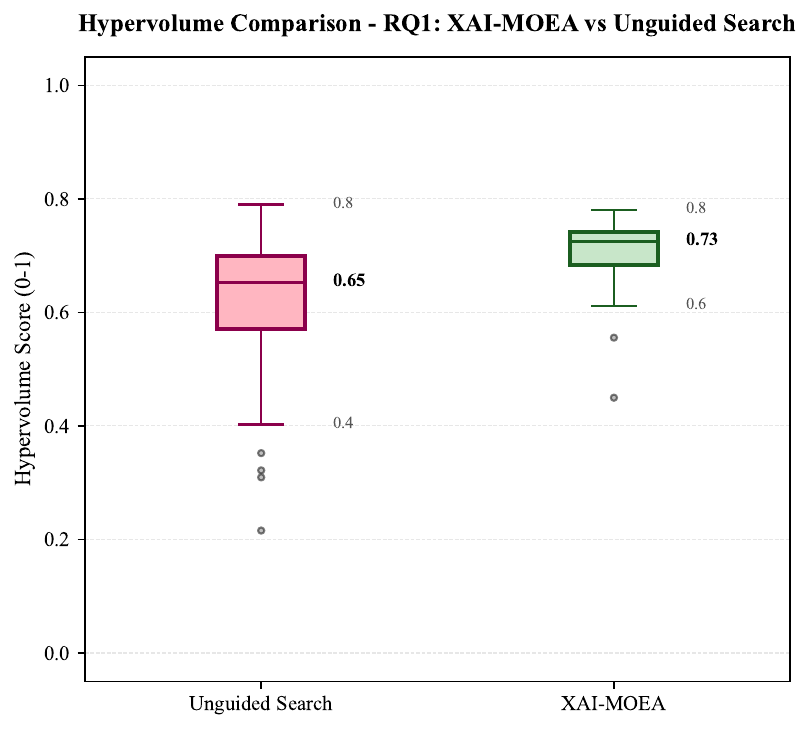}
    \caption{Hypervolume distribution comparison between XAI-MOEA and Unguided Search over 60 independent runs.}
    \label{fig:rq1_hv}
\end{figure}
\begin{figure} [ht]
    \centering    
    \includegraphics[width=0.70\linewidth]{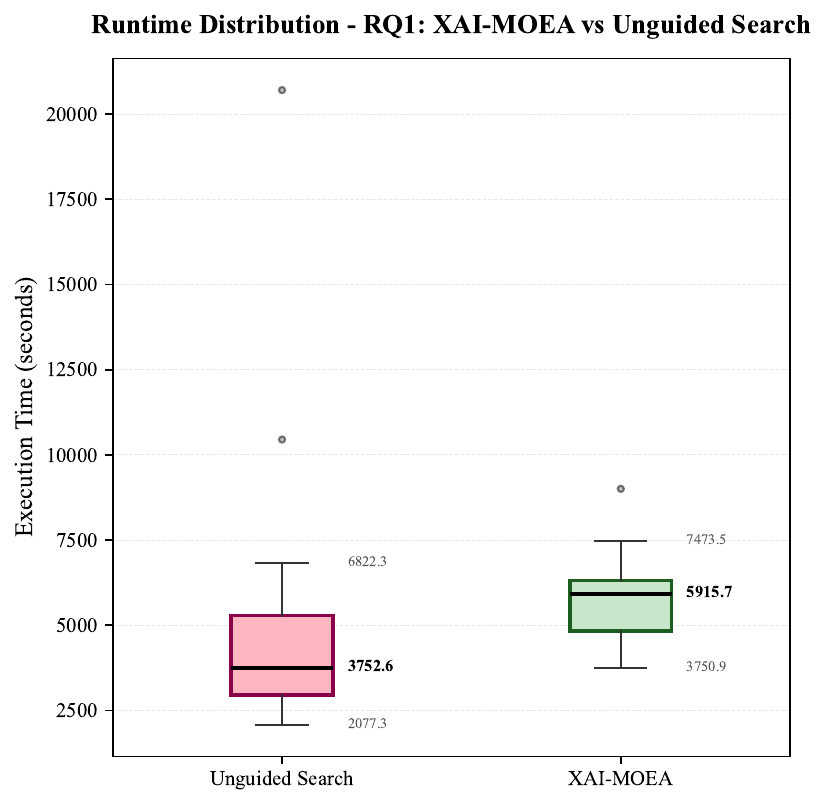}
    \caption{Runtime comparison between XAI-MOEA and Unguided Search during perturbation generation.}
    \label{fig:Runtime_XAI_VS_UNGUIDED}
\end{figure}
\textcolor{black}{The improvement was also observed in terms of Pareto-front quality, as depicted in the boxplots from Figure \ref{fig:rq1_hv}. Our approach obtained a median hypervolume of 0.73, whereas the hypervolume of the baseline dropped to 0.65. It is important to note that, within the hypervolume, we only consider those perturbations from the archive that were successful.}
\textcolor{black}{This metric summarizes the trade-off between perturbation cost (i.e., number of pixels changed) and failure-inducing effectiveness (i.e., time taken by the robot to misbehave). Subsequently, these results show that for those successful perturbations, overall, our approach generated higher-quality perturbation archives. The difference in hypervolume was also statistically significant after Bonferroni correction ($p < 0.00001$). The Vargha \& Delaney effect size for Unguided Search vs our approach was $\hat{A}_{12}=0.2017$, indicating that the proposed approach outperformed the baseline with a large effect size.}

\textcolor{black}{When considering computational time (Figure \ref{fig:Runtime_XAI_VS_UNGUIDED}), our approach approach required additional computation. The median runtime of our approach was approximately 5,915 seconds, compared with the 3,752 seconds for the baseline. This increase was expected because XAI-MOEA performed guided mutation and maintained an online Pareto archive during the search. However, the runtime remained within a practical range: all experiments were executed on a CPU-based personal laptop, and each independent run completed within approximately 1.6 hours on median. Therefore, although the proposed approach was computationally more expensive than unguided search, the additional cost was acceptable for an offline robustness-testing process and was compensated by substantially higher success rates and improved hypervolume.}

In summary, we can answer the first RQ as follows: 
\begin{center}
\fbox{
\begin{minipage}{0.94\linewidth}
\textbf{Answer to RQ1.} The proposed approach significantly outperformed the unguided baseline. It achieved a significantly higher success rate (70.0\% vs. 53.85\%) and a significantly higher hypervolume (0.73 vs. 0.65, $p < 0.00001$, $\hat{A}_{12}=0.2017$), while maintaining the runtime practical for offline robustness testing, with each run completing in approximately 1.6 hours.
\end{minipage}
}\end{center}
\subsection{RQ2 -- Ablation of the Optimization Formulation}
\textcolor{black}{RQ2 analyzed how the optimization formulation affected the reliability and quality of the generated perturbations. We compared three configurations: Variant I, which optimized perturbations for a single image; Variant II, which optimized perturbations across multiple representative images without XAI guidance; and our approach, which combines multi-image optimization with XAI-guided mutation. All configurations used the same perturbation representation, objective functions, and search budget. Therefore, this RQ isolated the effect of moving from single-image to multi-image optimization, and then investigates the effect of the XAI guidance.}
\begin{figure}[ht]
    \centering
    \includegraphics[width=0.65\linewidth]{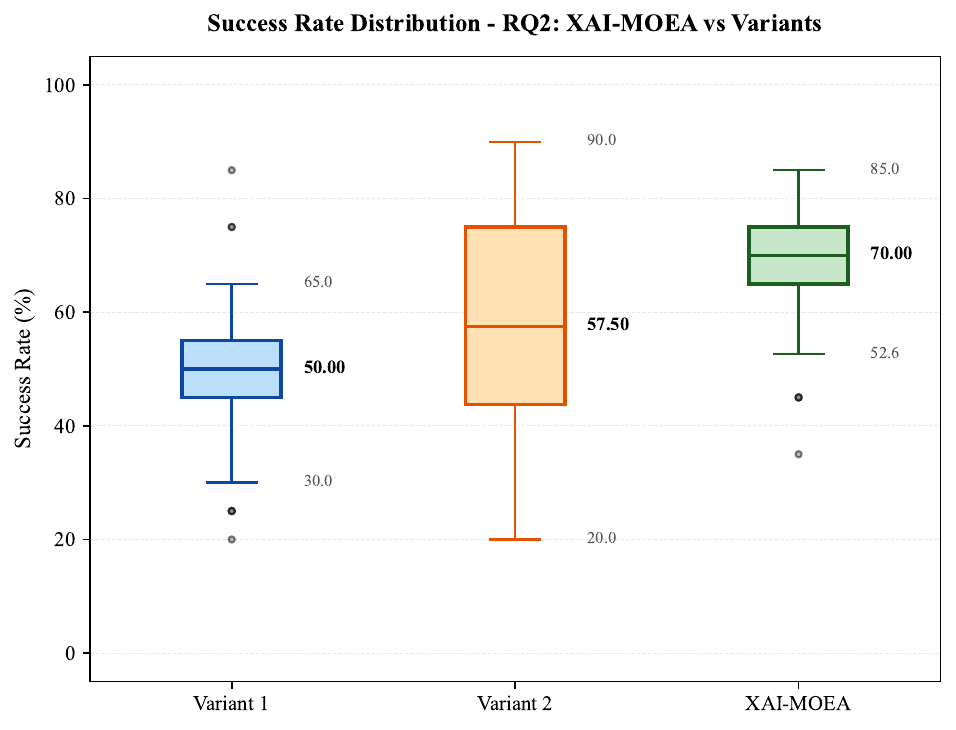}
    \caption{Success Rate comparison among XAI-MOEA and optimization variants.}
    \label{fig:rq2_sr}
\end{figure}
\begin{figure}[ht]]
\centering
\includegraphics[width=0.65\linewidth]{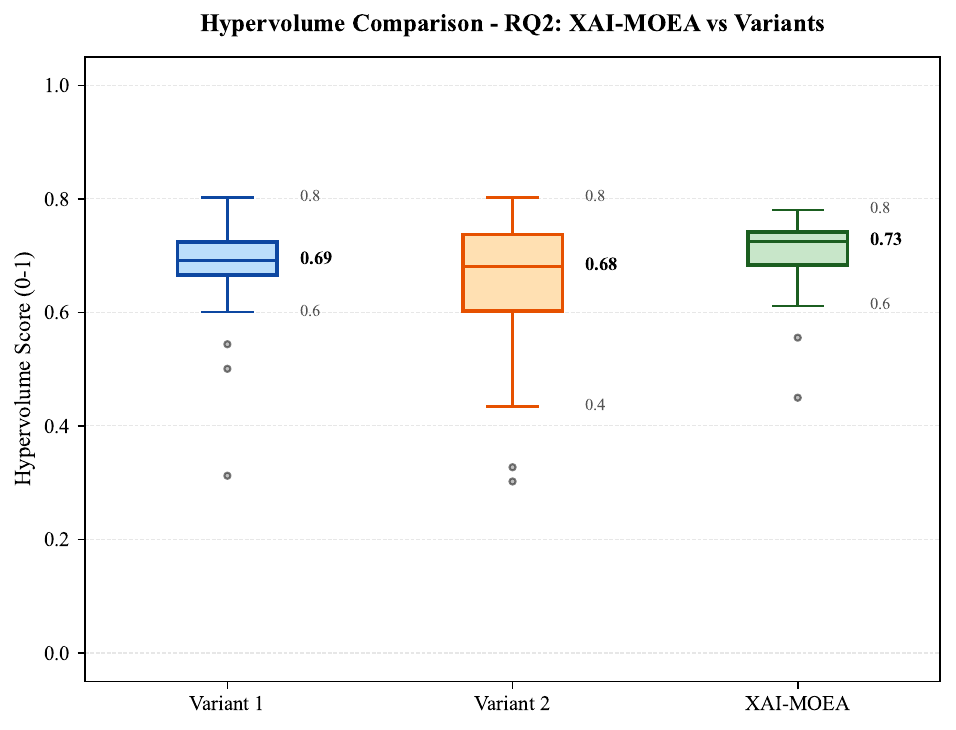}
\caption{Hypervolume comparison among XAI-MOEA and optimization variants.}
\label{fig:rq2_hv}
\end{figure}
\begin{figure}[ht]]
    \centering
    \includegraphics[width=0.65\linewidth]{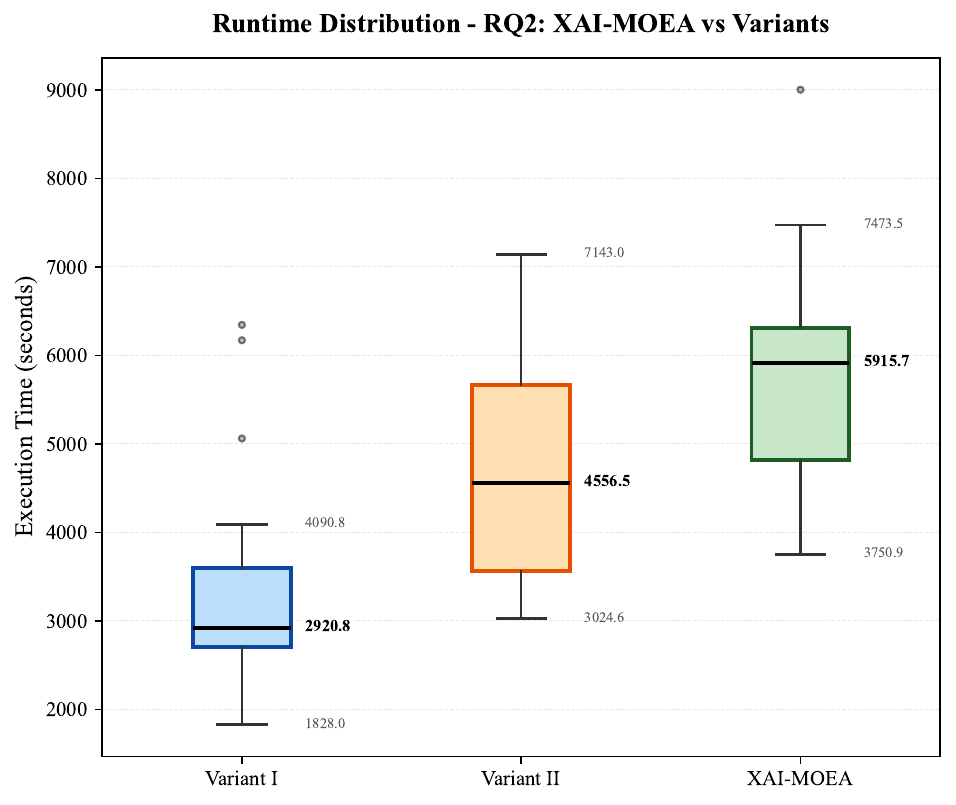}
    \caption{Runtime comparison among XAI-MOEA and optimization variants.}
    \label{fig:Runtime ea comparison}
\end{figure}
\textcolor{black}{The results showed a clear improvement in the failure-inducing capability of the generated perturbations. As shown in Figure~\ref{fig:rq2_sr}, Variant I achieved a median success rate of 50.0\%, while Variant II increased this value to 57.5\%. This suggests that evaluating candidate perturbations across multiple representative images helped generate perturbations that generalized better across operational scenarios. Our approach achieved the highest median success rate, reaching 70.0\%. This indicates that XAI-guided mutation provided an additional improvement over multi-image optimization alone in terms of perturbation success. The statistical analysis (\textcolor{black}{Table \ref{tab:stat_results}}) confirmed this trend: the difference between Variant I and our approach was statistically significant after Bonferroni correction ($p < 0.00001$), as was the difference between Variant II and our approach ($p < 0.001$). The corresponding Vargha \& Delaney effect sizes were $\hat{A}_{12}=0.1176$ for Variant I vs. our approach and $\hat{A}_{12}=0.3226$ for Variant II vs. our approach. Given the comparison direction, these values indicated that our approach obtained higher success rates in the majority of pairwise comparisons. In contrast, the difference between Variant I and Variant II was weaker and did not remain significant after correction ($p=0.05$).}

\textcolor{black}{When considering the hypervolume, the results showed a similar, although less statistically conclusive, trend. As shown in Figure~\ref{fig:rq2_hv}, our approach achieved the highest median hypervolume value, with 0.73, compared with 0.69 for Variant I and 0.68 for Variant II. Since hypervolume was computed only over successful perturbations and summarized the trade-off between perturbation cost and failure-inducing effectiveness, these results indicated that our approach, besides successful perturbations, it also tended to produce higher-quality successful perturbation archives. However, unlike success rate, the hypervolume differences did not remain statistically significant, as shown in Table \textcolor{black}{\ref{tab:stat_results}}. Therefore, these results should be interpreted as evidence of a positive trend rather than as a statistically confirmed improvement in Pareto-front quality.}

\textcolor{black}{The computational-time analysis showed the expected cost of each formulation. As shown in Figure~\ref{fig:Runtime ea comparison}, Variant I achieved the lowest median runtime, with approximately 2920 seconds per run. Variant II required approximately 4556 seconds, reflecting the additional cost of evaluating each candidate perturbation across multiple representative images. Our approach required the highest median runtime, approximately 5915 seconds, due to the combination of multi-image evaluation, XAI-guided mutation, and archive maintenance. Nevertheless, this cost remained practical for offline robustness testing: even the most expensive configuration completed in approximately 1.6 hours on median on a CPU-based laptop.}

In summary, we can answer RQ2 as follows:
\begin{center}
\fbox{
\begin{minipage}{0.94\linewidth}
\textbf{Answer to RQ2.} The optimization formulation had a clear impact on perturbation reliability. Moving from single-image optimization to multi-image optimization improved the median success rate from 50.0\% to 57.5\%, and adding XAI-guided mutation further increased it to 70.0\%. The improvement of our approach over both variants was statistically significant for success rate. Our approach also achieved the highest median hypervolume (0.73), although the hypervolume differences did not remain significant after Bonferroni correction. This improvement came with higher runtime, but the median execution time remained practical for offline robustness testing.
\end{minipage}
}
\end{center}

\begin{table}[t]
\centering
\caption{Mann--Whitney \(U\) test results for success rate and hypervolume.}
\label{tab:stat_results}

\begin{threeparttable}
\small
\setlength{\tabcolsep}{4pt}
\renewcommand{\arraystretch}{1.08}

\begin{tabularx}{\columnwidth}{
    @{}
    l
    >{\raggedright\arraybackslash}X
    S[table-format=1.4]
    S[table-format=<1.5]
    S[table-format=1.4]
    S[table-format=<1.5]
    @{}
}
\toprule
& 
& \multicolumn{2}{c}{Success rate}
& \multicolumn{2}{c}{Hypervolume} \\
\cmidrule(lr){3-4}
\cmidrule(l){5-6}
RQ & Comparison
   & {$\hat{A}_{12}$} & {$p$-value}
   & {$\hat{A}_{12}$} & {$p$-value} \\
\midrule

RQ1
& Unguided search vs.\ Variant 1
& 0.5521 & 0.325
& 0.3086 & <0.001 \\

& Unguided search vs.\ Variant 2
& 0.4283 & 0.176
& 0.4006 & 0.060 \\

& Unguided search vs.\ XAI-MOEA
& 0.2222 & <0.001
& 0.2017 & <0.001 \\

\addlinespace[3pt]

RQ2
& Variant 1 vs.\ Variant 2
& 0.3711 & 0.050
& 0.5697 & 0.189 \\

& Variant 1 vs.\ XAI-MOEA
& 0.1176 & <0.001
& 0.3800 & 0.050 \\

& Variant 2 vs.\ XAI-MOEA
& 0.3226 & <0.001
& 0.3572 & 0.010 \\

\bottomrule
\end{tabularx}

\begin{tablenotes}[flushleft]
\footnotesize
\item \(\hat{A}_{12}>0.5\) favors the first algorithm in the comparison,
whereas \(\hat{A}_{12}<0.5\) favors the second. Statistical significance is
assessed using a Bonferroni-corrected threshold of \(\alpha=0.0083\).
\end{tablenotes}
\end{threeparttable}
\end{table}

\subsection{RQ3 -- Sim-to-Real Transfer:}

\textcolor{black}{RQ3 evaluated whether perturbations generated and evaluated in simulation also transferred to the physical LeoRover. This analysis was different from RQ1 and RQ2. Instead of comparing different optimization configurations, we assessed whether the Gazebo simulator provided a useful approximation of physical robot behavior for the perturbations generated by our approach. This validation was necessary because executing the full search directly on the physical robot would have been impractical due to hardware resets, battery monitoring, manual setup, and execution time. Therefore, simulation was used as the main evaluation environment, while a stratified subset of perturbations was selected for physical validation.}

\begin{figure}[ht]
    \centering
    \includegraphics[width=0.65\linewidth]{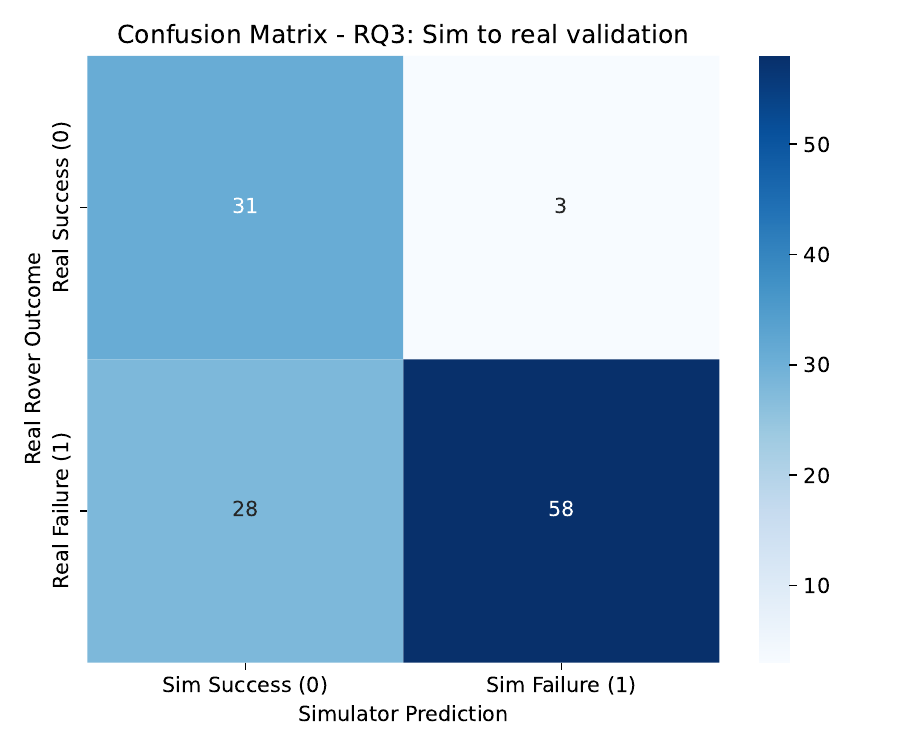}
    \caption{Sim-to-real consistency analysis based on binary execution outcomes.}
    \label{fig:Confusion_matrix}
\end{figure}

\textcolor{black}{We first analyzed the consistency between simulation and physical execution as a binary outcome: whether a perturbation induced an out-of-bounds failure or not. Figure~\ref{fig:Confusion_matrix} shows the resulting confusion matrix. Since each perturbation was executed twice on the simulator, we treated each simulator run as an independent prediction, resulting in 120 total comparisons (60 perturbations × 2 runs). Out of these 120 predictions, 58 were true positives (simulator predicted failure and physical robot failed), 3 were false positives (simulator predicted failure but physical robot was safe), 28 were false negatives (simulator predicted safety but physical robot failed), and 31 were true negatives (simulator predicted safety and physical robot was safe). These results showed that a substantial part of the simulator outcomes was consistent with physical execution.}

\begin{table}[ht]
\centering
\small
\caption{Derived Metrics for Sim-to-Real Outcome Consistency.}
\label{tab:rq3_metrics}
\begin{tabular}{l r p{6cm}}
\hline
\textbf{Metric} & \textbf{Value} & \textbf{Interpretation} \\ 
\hline

False Positive Rate ($\text{FPR}$) 
& $0.0882$ 
& Proportion of perturbation structures predicted as failures in simulation but remaining nominal on the physical platform. \\

Precision 
& $0.9508$ 
& Probability that a perturbation structure identified as producing failure in simulation also produces failure physically. \\

Recall (True Positive Rate) 
& $0.6744$ 
& Proportion of physical failures detected during simulation evaluation. \\

$\text{F}_1$-Score 
& $0.7891$ 
& Overall balance between Precision and Recall. \\

\hline
\end{tabular}
\end{table}

\textcolor{black}{The most relevant result was the high precision obtained by the simulator. The \textcolor{black}{precision was 0.9508}, meaning that perturbations identified as failure-inducing in simulation were very likely to also produce failures on the physical robot. Only three perturbations were false positives, i.e., predicted as failures in simulation but not reproduced as failures on hardware. This indicated that simulation was effective as a pre-screening mechanism for identifying perturbations that were likely to expose real system-level failures. In practical terms, this means that simulation could substantially reduce the number of perturbations that need to be executed on the physical robot, while still prioritizing candidates with a high probability of transferring to hardware.}

\textcolor{black}{However, the results also showed that simulation was not sufficient to rule out physical failures. \textcolor{black}{ The recall was 0.6744, and 28 perturbations were false negatives}: they appeared benign in simulation but produced out-of-bounds failures on the physical robot. This indicated that the sim-to-real gap remained relevant. These false negatives may be explained by differences between the simulated and physical environments, such as sensor noise, robot dynamics, friction, illumination, camera behavior, or other unmodeled physical effects. Therefore, while simulation was useful for finding real failure-inducing perturbations, according to our results, it should not be interpreted as proof that a perturbation is safe when no failure is observed in simulation.}

\begin{figure}
    \centering
    \includegraphics[width=0.65\linewidth]{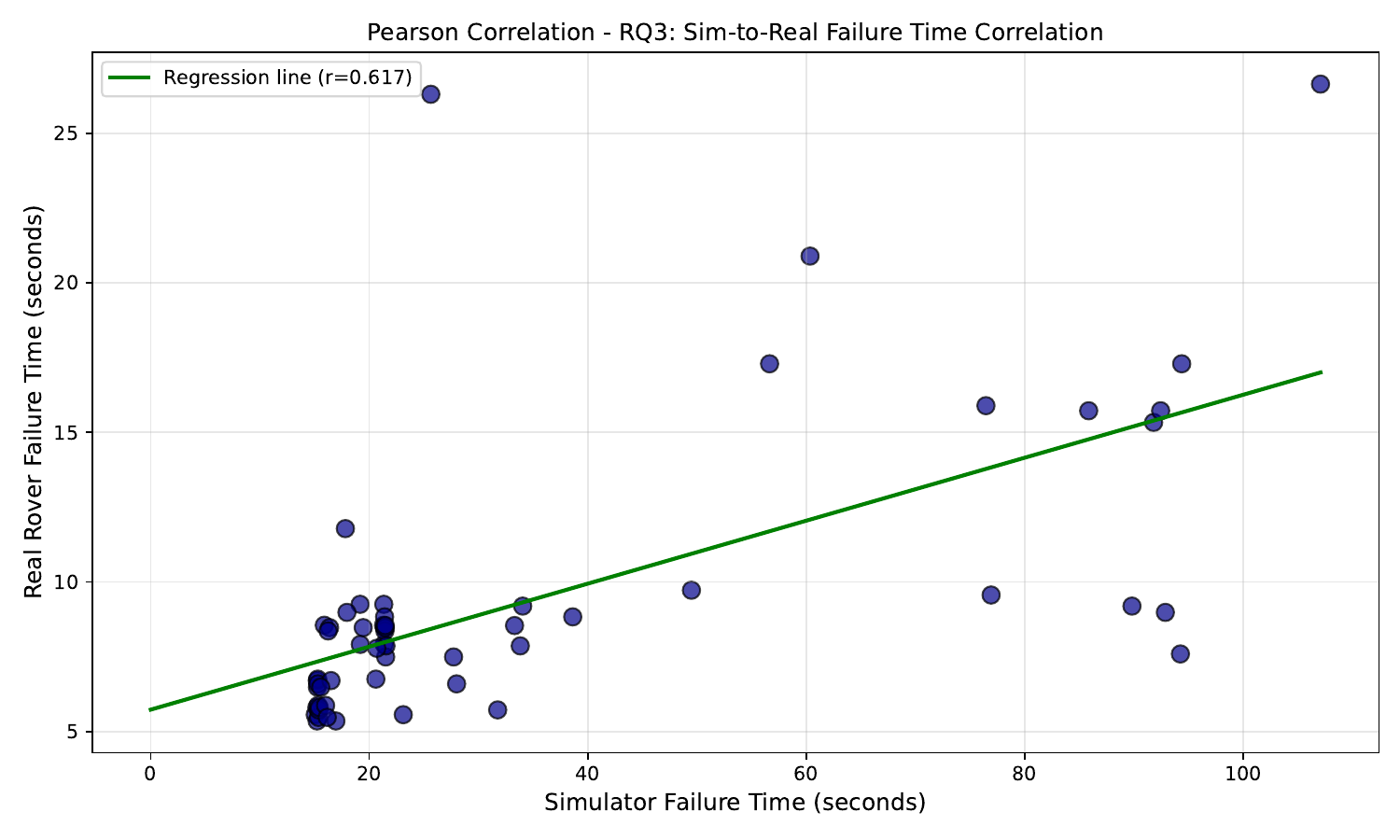}
    \caption{Correlation between simulated and physical failure times for transferred perturbation structures.}
    \label{fig:pearson_corelation}
\end{figure}

\textcolor{black}{Beyond the binary outcome, we also analyzed whether simulation preserved the relative severity of transferred failures. For the 27 perturbations that produced out-of-bounds failures in both environments, we compared the simulated and physical failure times. As shown in Figure~\ref{fig:pearson_corelation}, the Pearson correlation between simulated and physical failure time was strong and statistically significant \textcolor{black}{($r=0.6170$, $p=2.5 \times 10^{-7}$)}. This indicated that simulation preserved the relative ranking of perturbation severity: perturbations that failed earlier in simulation also tended to fail earlier on the physical robot. However, the absolute failure times were not identical. In particular, physical executions generally reached the out-of-bounds condition earlier than simulated executions. This is attributed to the different circuit of the physical test bench and the simulator. \textcolor{black}{
This discrepancy in absolute failure times is explained by the different circuit geometries used in each environment. The simulated track was a large oval with long straight segments, while the physical test bench was a closed-loop U-shaped track with tighter turns and shorter straight sections. On the U-shaped track, the robot must change direction more frequently and operates closer to the boundaries for a greater proportion of its trajectory. Therefore, when a perturbation induces unstable behavior, the physical robot reaches the out-of-bounds condition earlier than in simulation not because of any mismatch in dynamics or timing, but because the physical circuit geometrically constrains the robot's recovery space more severely. We selected two different circuit shapes to evaluate whether our approach generalizes beyond a single track layout, not to match absolute failure times. Consequently, the absolute failure times are not directly comparable, but the strong correlation in relative severity (r = 0.6170) confirms that simulation preserves the ordering of perturbation risk across different track geometries.
}}

\textcolor{black}{In summary, RQ3 can be summarized as follows:}

\begin{center}
\fbox{
\begin{minipage}{0.94\linewidth}
\textbf{Answer to RQ3.} Simulation provided a useful but imperfect approximation of physical robot behavior. It achieved high precision (0.95), showing that perturbations identified as failure-inducing in simulation were likely to transfer to the physical LeoRover. However, the lower recall (0.67) and the presence of 28 false negatives showed that simulation could miss physical failures. For perturbations that failed in both environments, simulation preserved the relative severity trend ($r=0.6170$, $p=2.5 \times 10^{-7}$), although physical failures generally occurred earlier. Therefore, simulation was effective for pre-screening and prioritizing perturbations, but physical validation remains necessary for safety-critical robustness assessment.
\end{minipage}
}
\end{center}

\section{Threats to Validity}

We now discuss the threats of our study and the actions we took to try to mitigate them.

\textbf{External validity:} Our evaluation considers one DNN controller, one robotic platform, and one navigation task, which limits the generalization of the results to other models, robots, and operational environments. We partially mitigate this threat by evaluating the perturbations in both simulation and the physical system.

\textbf{Internal validity:} The selected clustering, attribution, perturbation, and search parameters may influence the results. We used the same perturbation representation, objectives, and search budget across the compared configurations, and conducted an ablation study to isolate the effects of multi-image optimization and XAI guidance. During physical executions, we monitored the battery and verified nominal robot behavior before continuing the experiments.

\textbf{Conclusion validity:} The stochastic nature of the search and simulation may affect the observed results. We mitigated this threat by performing 30 independent search runs, replaying each archive twice in simulation, and applying statistical tests with correction for multiple comparisons and effect-size analysis. 

\textbf{Construct validity:} DNN-output deviation is used as a proxy objective during search and does not necessarily imply a system-level failure. We therefore evaluate the generated perturbations using multiple metrics, including actual out-of-bounds executions, success rate, failure time, perturbation cost, and hypervolume.

\section{Related Work}

\textcolor{black}{Adversarial robustness has been extensively studied in the context of generic computer-vision tasks, particularly image classification and object detection. Early work showed that small, carefully crafted perturbations can cause deep neural networks to misclassify otherwise correctly classified inputs~\cite{szegedy2013intriguing,goodfellow2015explaining,carlini2017towards}. Subsequent studies extended this observation in several directions, including universal perturbations that generalize across many inputs~\cite{moosavi2017universal}, sparse perturbations generated through evolutionary search~\cite{su2019one,alzantot2019genattack}, and adversarial patches or physical-world attacks targeting classifiers and object detectors~\cite{brown2017adversarial,eykholt2018robust,liu2018dpatch}. More recent work has also explored the use of saliency, attention, or attribution methods to focus adversarial perturbations on influential image regions~\cite{wicker2018feature,wei2022towards,dai2023saliency}. These studies are important because they establish the vulnerability of deep vision models and provide a rich set of perturbation-generation mechanisms. However, most of them evaluate attacks at the model level, using classification accuracy, detection confidence, or label changes as the main outcome. In contrast, our work focuses, on the one hand, on robustness testing rather than attack, and on the other hand, targeting DNN-based robotic navigation, where the relevant question is not only whether the DNN output changes, but whether the perturbation propagates through the closed-loop control pipeline and induces unsafe system-level behavior, such as the robot leaving its intended track.}

\textcolor{black}{A second line of work has focused on testing DNN-based cyber-physical and robotic systems at the scenario or environment level. For autonomous driving, DeepTest applies metamorphic transformations such as rain, fog, brightness changes, and affine image transformations to expose erroneous driving behaviors~\cite{tian2018deeptest}, while DeepRoad uses GAN-based image-to-image translation to generate realistic weather variations~\cite{zhang2018deeproad}. Other approaches move further toward search-based exploration of the operational domain. DeepJanus explores the frontier between safe and unsafe behaviours of autonomous systems by manipulating high-level scenario parameters~\cite{riccio2020model}, whereas DeepHyperion uses illumination search to explore feature spaces and generate diverse failure-inducing inputs~\cite{zohdinasab2021deephyperion}. These approaches are highly relevant because they evaluate DNNs in CPS and robotics contexts and consider system-level failures rather than only model-level errors. Nevertheless, their perturbations are typically defined as semantic, environmental, or scenario-level changes, such as road geometry, weather, lighting, or other high-level scene properties. Our approach differs by keeping the operational scenario fixed and searching directly in the camera-input space for localized perturbation structures, therefore focusing on testing the robustness of robotic DNN controllers. Rather than generating new scenes or modifying simulator parameters, we generate sparse image perturbations that can be applied consistently across representative camera images and then evaluate whether they induce failures in the robotic navigation system.}

\textcolor{black}{Closer to our work are studies that perturb the camera input of autonomous or robotic systems. DeepBillboard generates physical billboard perturbations that affect the steering decisions of autonomous-driving systems across multiple frames~\cite{zhou2020deepbillboard}. Other studies investigate image-specific and image-agnostic perturbations against end-to-end autonomous driving models~\cite{boloor2019simple}. Other works investigate adaptive image perturbations for vision-guided vehicles~\cite{yoon2023learning}, physical patches for monocular depth estimation~\cite{cheng2022physical}, or broader perturbation-based testing frameworks for advanced driver-assistance systems~\cite{leonhard2026perturbationdrive}. These studies show that perturbing camera images can affect autonomous perception and control. However, most existing approaches either optimize perturbations for individual inputs, rely on gradient-based or task-specific formulations, use predefined perturbation operators, or do not explicitly exploit explainability to guide the search. Our work differs by combining several elements that have mostly been studied separately: (1) we first select representative operational images to avoid overfitting to a single frame; (2) we aggregate Integrated Gradients over these images to obtain a global attribution map; (3) we use this map as a probabilistic prior to guide mutation in a multi-objective evolutionary search; and (4) we generate sparse universal perturbation structures that balance perturbation impact and cost. Finally, unlike studies that remain at the model or simulation level, we evaluate the generated perturbations both in simulator and on a physical robot, allowing us to assess their system-level effect and their sim-to-real transferability.}

\section{Conclusion}

Robustness testing of DNN-based autonomous robots remains challenging because perturbations optimized for individual images may overfit.
To address this problem, we proposed an XAI-guided multi-objective evolutionary approach that generates sparse perturbations across a compact set of representative images and guides the search using a global Integrated Gradients attribution map.
The proposed approach achieved a median success rate of 70.0\%, compared with 53.85\% for baseline techniques, and improved the median hypervolume from 0.65 to 0.73.
The ablation study showed that multi-image optimization increased the median success rate from 50.0\% to 57.5\%, while incorporating XAI guidance further increased it to 70.0\%.
The improvement of the proposed approach over the single-image and unguided multi-image variants was statistically significant for success rate, although the hypervolume differences among the evolutionary variants were not significant.
These improvements required additional computation, but the median execution time of approximately 1.6 hours is practical for offline robustness testing.
The physical evaluation showed that simulation was effective at identifying transferable failures, reaching a precision of 0.95, which means that most of the perturbations are successful in the real physical world. However, its recall decreased to 0.67, which indicates that some perturbations that are not successful in simulation may still lead to failures on the real robot.
Overall, the results show that XAI-guided multi-image search improves the generation of effective perturbation-based tests, while targeted physical validation remains necessary.
Future work will evaluate the approach on additional robots, controllers, environments, attribution methods, and perturbation types.

\section*{Replication Package}
The replication package is publicly available on Zenodo at
\url{https://doi.org/10.5281/zenodo.21496097}. The virtual machine is
available separately at \url{https://doi.org/10.5281/zenodo.21497744}.
These are provided as separate Zenodo records due to the large size of
the virtual machine image.

\section*{Declaration of Competing Interests}

The authors declare the following financial interests/personal relationships which may be considered as potential competing interests: Aitor Arrieta reports financial support was provided by University of Mondragon. If there are other authors, they declare that they have no known competing financial interests or personal relationships that could have appeared to influence the work reported in this paper.

\section*{Acknowledgments}
This work is supported by the InnoGuard Marie Skłodowska-Curie Doctoral Network (Grant Agreement No. 101169233). This work has also been partly funded by the Spanish Ministry of Science, Innovation and Universities (project PID2023-152979OA-I00), funded by MCIU /AEI /10.13039/ 501100011033/FEDER, UE). The authors are part of the Software and Systems Engineering research group of Mondragon Unibertsitatea (IT1519-22), supported by the Department of Education, Universities and Research of the
Basque Country.

\bibliographystyle{unsrt}  


\bibliography{sources}

\end{document}